%% file: conference_071817.tex
\documentclass[conference]{IEEEtran}
\IEEEoverridecommandlockouts
\usepackage{cite}
\usepackage{amsmath,amssymb,amsfonts}
\usepackage{algorithmic}
\usepackage{graphicx}
\usepackage{textcomp}
\usepackage[hidelinks]{hyperref}

\usepackage{aveasmargins}

\usepackage{overpic}
\usepackage[font=footnotesize]{subcaption}
\usepackage{tikz}
\usepackage{units}
\usepackage{txfonts}
\let\mathbb=\varmathbb

\include{includes}

\def\BibTeX{{\rm B\kern-.05em{\sc i\kern-.025em b}\kern-.08em
    T\kern-.1667em\lower.7ex\hbox{E}\kern-.125emX}}
\begin{document}

\title{Accuracy Evaluation of a Lightweight Analytic Vehicle Dynamics Model
for Maneuver Planning*
\thanks{* This publication was written in the framework of KAMO: Karlsruhe Mobility / Profilregion Mobilit\"atssysteme Karlsruhe (\href{https://www.kamo.one}{kamo.one}), which is funded by the Ministry of Science, Research and the Arts and the Ministry of Economic Affairs, Labour and Housing in Baden-W\"urttemberg and as a national High Performance Center by the Fraunhofer-Gesellschaft.}
}

\author{\IEEEauthorblockN{J.R. Ziehn}
\IEEEauthorblockA{\small \textit{Dept. MRD} \\
\textit{Fraunhofer IOSB}\\
Karlsruhe, Germany\\
jens.ziehn@iosb.fraunhofer.de}
\and
\IEEEauthorblockN{M. Ruf}
\IEEEauthorblockA{\small \textit{Dept. MRD} \\
\textit{Fraunhofer IOSB}\\
Karlsruhe, Germany\\
\small miriam.ruf@iosb.fraunhofer.de}
\and
\IEEEauthorblockN{M. Roschani}
\IEEEauthorblockA{\small \textit{Dept. MRD} \\
\textit{Fraunhofer IOSB}\\
Karlsruhe, Germany\\
\small masoud.roschani@iosb.fraunhofer.de}
\and
\IEEEauthorblockN{J. Beyerer}
\IEEEauthorblockA{\small \textit{Fraunhofer IOSB}\\
\textit{\& Vision and Fusion Lab,}\\\textit{Karlsruhe Institute of Technology KIT}\\
Karlsruhe, Germany}
%\and
%\and
%\and
}

\maketitle

\begin{abstract}
Models for vehicle dynamics play an important role in maneuver planning for automated driving. They are used to derive trajectories from given control inputs, or to evaluate a given trajectory in terms of constraint violation or optimality criteria such as safety, comfort or ecology. Depending on the computation process, models with different assumptions and levels of detail are used; since maneuver planning usually has strong requirements for computation speed at a potentially high number of trajectory evaluations per planning cycle, most of the applied models aim to reduce complexity by implicitly or explicitly introducing simplifying assumptions. While evaluations show that these assumptions may be sufficiently valid under typical conditions, their effect has yet to be studied conclusively.

We propose a model for vehicle dynamics that is convenient for maneuver planning by supporting both an analytic approach of extracting parameters from a given trajectory, \emph{and} a generative approach of establishing a trajectory from given control inputs. Both applications of the model are evaluated in real-world test drives under dynamic conditions, both on a closed-off test track and on public roads, and effects arising from the simplifying assumptions are analyzed.
\end{abstract}

\begin{IEEEkeywords}
Vehicle dynamics, vehicle models, automated driving, trajectory planning
\end{IEEEkeywords}

\aveasSetMargins{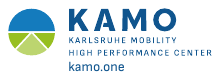}
\aveasSetIEEEFoot{2020}
\aveasSetIEEEHead{%
J. R. Ziehn, M. Ruf, M. Roschani and J. Beyerer, "Accuracy Evaluation of a Lightweight Analytic Vehicle Dynamics Model for Maneuver Planning," 2020 5th International Conference on Robotics and Automation Engineering (ICRAE), Singapore, Singapore, 2020, pp. 197--205
}{10.1109/ICRAE50850.2020.9310898}

%%%%%%%%%%%%%%%%%%%%%%%%%%%%%%%%%%%%%%%%%%%%%%%%%%%%%%%%%%%%%%%%%%%%%%%%%%%%%%%%
\section{INTRODUCTION AND MOTIVATION}

%\the\columnwidth
%
%\the\textheight
%
%\the\textwidth

Models of vehicle dynamics in the domain of automated driving serve various different purposes: They are used for deriving parameters of control systems, for generating simulations, for aiding environment perception by estimating and tracking other vehicles' parameters, and within the task of trajectory planning. Depending on the purpose, a minimum level of detail may be required, and/or a maximum computational effort. In this paper, we focus mainly on the purpose of trajectory planning (with other possible applications indicated), understood here as the task of finding a \emph{trajectory} for the ego vehicle
\begin{equation}
	\traj: [t\subs{s}, t\subs{e}] \rightarrow \mathbb R^2 \;\;\text{with}\;\;  \traj(t) = \begin{bmatrix} x(t)\\ y(t)\end{bmatrix}, \label{eq:traj}% \;\;\text{and}\;\; x,y \in C^2,
\end{equation}
as the Euclidean surface coordinates of the center of the rear (or \emph{unsteered}) axle over a limited time interval (usually several seconds into the future). Various processes are used by which such a trajectory is chosen; arguably, the most common are those that can be expressed as an optimization process which evaluates a given trajectory $\traj \in \Xi$ by a cost measure $\mathcal C : \Xi \rightarrow \mathbb R$ (a \emph{functional}), and aims to find a trajectory to minimize $\mathcal C[\traj]$ (e.g. \cite{ulbrich, kasac_deur, paden, walting2}). Vehicle dynamics are typically relevant to evaluating $\mathcal C$, e.g. for determining whether a trajectory violates physical constraints, or for rating qualities of a trajectory such as comfort, safety, etc. (cf. \cite{rosolia, rucco}).

We distinguish (cf. also \cite{gonzalez2015review}) between \emph{iterative} approaches that take a candidate trajectory $\traj^1$, and use it to seek a next trajectory $\traj^2$ such that $\mathcal C[\traj^2] < \mathcal C[\traj^1]$; and \emph{generative} processes that build a candidate trajectory from scratch, often by combinatorial and/or graph-based approaches (e.g. \cite{AJANOVIC2018255, zhao_wang_heuristic_programming, brechtel_gindele_dillmann, ferguson2008motion, pivtoraiko2005efficient}). A third category that exclusively compares few precomputed trajectories is disregarded because the properties described here are usually not relevant in this case.

Iterative processes often require derivatives of $\mathcal C$ w.r.t. $\traj$, and therein of $\traj$ w.r.t. $t$, so it is beneficial if the dynamics do not depend on high-order derivatives. An example is the Euler--Lagrange model of trajectory planning, where the cost functional is given by
\begin{equation}
	\mathcal C[\traj] = \int_{t\subs{s}}^{t\subs{e}}\!\!\dee t\, L(\traj(t), \dot\traj(t), \ddot\traj(t), \dotsc , \dee^n/(\dee t)^n \traj),
\label{eq:elm}
\end{equation}
and where $L$ (the \emph{Langrangian}) is a function that assigns local (i.e. at individual $t$) costs to be integrated to cumulative costs $\mathcal C$. In this model, iterative optimization is particularly efficient, but the effort scales with the order $n$ of derivatives required to evaluate $L$.

Models conforming to \eqref{eq:elm} also allow for transformation to a \emph{generative} model suitable for dynamic programming (cf. \cite{ZR.2015}). In this, a discretized search tree is generated over trajectory time $t$, which can be solved to obtain a globally optimal trajectory (given the discretization). This model, as well as similar models, builds a trajectory over time in the form of a differential equation, where it is beneficial to have a small state space describing each point of a trajectory, since the required computational time and space scales with the order $n$, possibly exponentially (cf. \cite{ZR.2015, Ruf2018_1000085281, huang_wu_dynamic_prog}). 

With trajectory planning being subject to strong realtime requirements, and typically requiring the evaluation of multiple trajectories per cycle at multiple cycles per second, the use of lightweight models is of critical importance; hence, simplified models of vehicle dynamics are used (also called \emph{kinematic models}) that disregard masses and forces, and primarily describe the vehicle by geometrical properties.

These models are efficiently computable, but the degree of accuracy that can be achieved under realistic conditions for automated driving requires careful evaluation. In \cite{icmc}, a model is described particularly from the optimization perspective of twice continuously differentiable trajectories (hence labeled $C^2$ model, used as a basis of this paper), but only evaluated in simulated scenarios, indicating a very good degree of accuracy. Variations of the model are used for automated maneuver planning e.g. in \cite{ziegler, RZ.2014a}. A similar model from the perspective of system control is given in \cite{kang_lee_chung} and compared to a true dynamic (i.e. physical) model at the purpose of lane keeping, in both simulated and real test drives. The authors conclude that the kinematic model is sufficient for the purpose, provided that there is no considerable loss in tire-road friction; a similar conclusion is given in \cite{kong2} for the application of model-predictive control (MPC). By comparing the performance of the kinematic model to other models, \cite{polack} finds that for limited accelerations (namely below $0.5\,\mu\,\unit{g}$, with $\mu$ being the friction coefficient) the kinematic model achieves extrapolation results that are similar to more complex vehicle models; however the properties of such models under typical driving conditions are yet to be thoroughly evaluated.

To contribute to this evaluation, this paper uses a model based on \cite{icmc}, whose aim is to establish the most relevant vehicle parameters in motion planning in relation to its twice-differentiable trajectory (instead of, e.g., control inputs), to enable its efficient use within direct trajectory optimization, as posed e.g. in \eqref{eq:elm}. It only uses explicit models and assumptions, and is chosen such that the model is both efficiently computable for analytic approaches (namely extracting these parameters from a given trajectory), and for generative approaches (namely establishing a trajectory e.g. from known control inputs; cf. Sec.~\ref{sec:forward-model}). Besides applications in trajectory planning, where a trajectory is usually known to model precision, the model can be used to extract parameters from uncertain observations of trajectories, e.g. in vehicle, roadside or airborne sensors, as well as in lightweight simulations.

After describing the model and its simplifying assumptions, we briefly summarize the key results of previous simulation experiments in \cite{icmc}, and evaluate the model performance in detail in real-world drives on a closed-off test track (Sec.~\ref{sec:tests-campusost}) and on public roads (Sec.~\ref{sec:tests-interchange}), to quantify resulting errors by trajectory parameters. Section~\ref{sec:conclusion} provides a conclusion and an outlook to future work.

\section{MODEL DESCRIPTION}\label{sec:model}

\begin{figure}%
\begin{overpic}[width=\columnwidth%,grid
			]{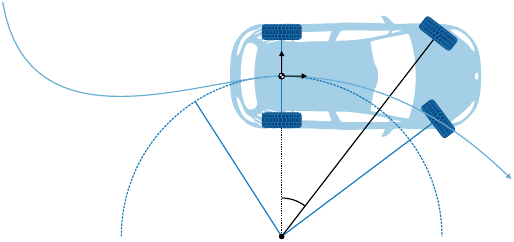}%
				\put(56,10){\footnotesize $\delta\subs{FL}$}
				\put(5,40){\footnotesize $\traj$}
				\put(50,29){\footnotesize $\traj(t)$}
				\put(52,35){\footnotesize $\boldsymbol N$}
				\put(58,33){\footnotesize $\boldsymbol T$}
				\put(43,8){\footnotesize $\displaystyle\frac{1}{\kappa}$}
				%\put(30,66.5){$r\subs{rear}$}
				%\put(89,58){$r\subs{front}$}
			\end{overpic}
\caption{Overview of the main geometric parameters in the $C^2$ model at a particular point $\traj(t)$ along the trajectory $\traj$, establishing the tangent $\boldsymbol T$ and normal $\boldsymbol N$ to derive i.a. the curvature $\kappa$ and the front left wheel angle $\delta\subs{FL}$.}%
\label{fig:c2golf}%
%\vspace{-15pt}
\end{figure}
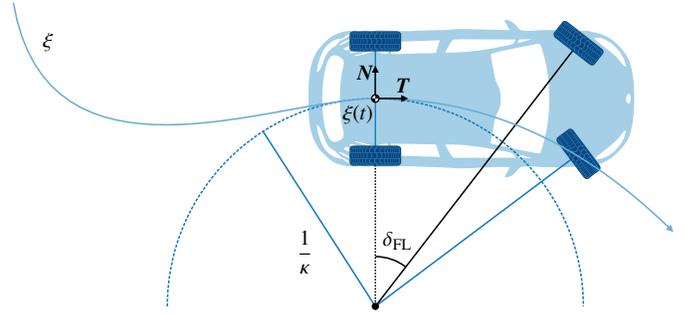

The evaluated model is based on a given vehicle's trajectory as in \eqref{eq:traj}, with the additional requirement that it be twice continuously differentiable, i.e. $x(t),y(t) \in C^2$. For cases where reverse gear driving may occur (e.g. parking scenarios), we additionally introduce a binary \emph{reverse gear} function
\begin{equation}
	R(t) = \begin{cases}\;1 & \text{while driving in reverse, and}\\\;0 & \text{otherwise.}\end{cases}
\end{equation}

\begin{figure}%
\begin{subfigure}{0.49\columnwidth}
\includegraphics{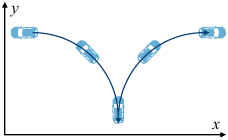}%
\caption{Path of the vehicle with a cusp}
\end{subfigure}
\hfill
\begin{subfigure}{0.49\columnwidth}
\includegraphics{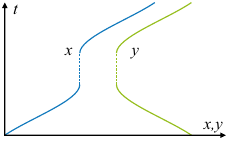}%
\caption{Smooth individual coordinates}
\end{subfigure}
\caption{A typical reversing trajectory where the \emph{path} (the trajectory parametrized by arc length) has a singularity, but the trajectory's coordinates are smooth since $\|\dot\traj\|=0$ at the cusp (dashed lines in (b)). In these cases, a consistent heading direction can be defined via \eqref{eq:tfill}.}%
\label{fig:cusp}%
\end{figure}

The vehicle's dynamics are then described with respect to the trajectory's tangential frame, whose basis are (in this order) the local tangent $\boldsymbol T$ (always pointing in the front direction of the car, cf. Fig.~\ref{fig:c2golf}) and the normal $\boldsymbol N$ given by
\begin{equation}
	\boldsymbol T(t) = (-1)^{R(t)} \frac{\dot\traj(t)}{\|\dot\traj(t)\|} \qquad\text{and} \qquad \boldsymbol N = \left[\begin{smallmatrix}0 & -1\\[2pt]1 &  \phantom{-}0\end{smallmatrix}\right] \boldsymbol T\label{eq:tangent}
\end{equation}
with a continuous (i.e. constant) $R(t)$ wherever the trajectory's derivative over time $\dot\traj(t)$ is non-vanishing. On any interval $\tau = [t_0, t_1]$ where $\|\dot\traj\| \equiv 0$ but
\begin{equation}
\boldsymbol 0 \neq  \lim_{t\nearrow t_0} \boldsymbol T(t) = \lim_{t\searrow t_1} \boldsymbol T(t) =: \boldsymbol T\subs{fill}\label{eq:tfill}
\end{equation}
we let $\boldsymbol T(t) \equiv \boldsymbol T\subs{fill}$. If, in a given trajectory, any such interval $\tau$ cannot be resolved this way, the trajectory changes directions without changing positions, thus the tangential space and the derived parameters cannot be uniquely determined at these times.\footnote{The trajectory may, however, still be completely driveable for vehicles that can turn around the center of their unsteered axle. For road vehicles however, this is an unusual design.} Otherwise, $\boldsymbol T(t)$ is thus defined and continuous everywhere in $(t\subs{s}, t\subs{e})$, even for trajectories containing stops and reversals whose \emph{path} (the trajectory parametrized by arc length) may not be continuously differentiable (cf. Fig.~\ref{fig:cusp}).

\pagebreak 
The model requires three main assumptions:
\begin{itemize}
	\item The $x$--$y$ surface has zero (Gaussian) curvature. Generalizations for curved surfaces are  not discussed here.
	\item The vehicle's wheels have neither \emph{lateral} nor \emph{longitudinal} slip, so that their direction and speed of rotation are always aligned with the motion of the wheel's frame. This in particular implies that the lateral speed (lateral slip) of the vehicle is zero, since the unsteered axle cannot move laterally, such that the vehicle's longitudinal \emph{x} axis (cf. Fig.~\ref{fig:c2golf}) always aligns with $\boldsymbol T$.
	\item The vehicle's roll and pitch angles are neglected; they are neither estimated, nor do they affect the modeled driving geometry.
\end{itemize}

With these assumptions, the lateral and longitudinal speeds and accelerations (w.r.t. the vehicle's coordinate system) can be established\footnote{Using the shorthand notation $\smash{\det\big[[\begin{smallmatrix}a \\b \end{smallmatrix}],[\begin{smallmatrix}c \\ d\end{smallmatrix}]\big] \!:=\!  (\big[\begin{smallmatrix}0 & -1\\1 & \phantom{-}0\end{smallmatrix}\big]\, \big[\begin{smallmatrix}a \\b \end{smallmatrix}\big])\transp \big[\begin{smallmatrix}c \\ d\end{smallmatrix}\big] \!=\! \det\big[\begin{smallmatrix}a & c\\b & d\end{smallmatrix}\big]}$} as
\begin{align}
	v\subs{lon} = \boldsymbol T \transp \dot\traj = (-1)^R \, \|\dot\traj\| \;&\text{,} \;\; v\subs{lat} = \boldsymbol N \transp \dot\traj \equiv 0\\[10pt]
	a\subs{lon} = \boldsymbol T\transp \ddot\traj = (-1)^R \, \frac{\dot\traj \transp \ddot\traj}{\|\dot\traj\|} \;&\text{,}\;\; a\subs{lat} = \boldsymbol N \transp \ddot\traj = (-1)^R \, \frac{\det[\dot\traj, \ddot\traj]}{\|\dot\traj\|}.
%\intertext{such that}
%\boldsymbol v = v\subs{lon}\, \boldsymbol T + v\subs{lat} \, \boldsymbol N \quad&\text{and}\quad \boldsymbol a = a\subs{lon}\, \boldsymbol T + a\subs{lat} \, \boldsymbol N,
\end{align}

To obtain the steering geometry, we compute the trajectory's local curvature (which is not necessarily continuous in the model) using
\begin{equation}
\begin{aligned}
	\kappa = \boldsymbol N \transp (\tfrac{\dee}{\dee s}\boldsymbol T)= \frac{\det[\boldsymbol T,\dot{\boldsymbol T}]}{\|\dot\traj\|}= (-1)^R \frac{\det[\dot\traj,\ddot{\traj}]}{\|\dot\traj\|^3},\label{eq:curvatureanalytic}
\end{aligned}
\end{equation}
which also provides a briefer form of the accelerations as
\begin{equation}
a\subs{lon} = \frac{\dee v\subs{lon}}{\dee t} \quad\text{and}\quad a\subs{lat} = \frac{\det[\dot\traj, \ddot\traj]}{\|\dot\traj\|} = \kappa \, v\subs{lon}^2.
\end{equation}

Since the curvature is the inverse of the vehicle's turning radius, $\kappa = 1/R$, all points on the chassis are known to rotate around the point $\traj(t) + \boldsymbol N / \kappa$ when $\kappa \neq 0$. Specifically for any wheel mounted at a point $[d\subs{lon}, d\subs{lat}]$ relative to the chassis' origin, it is aligned with its direction of movement iff it has a steering wheel angle of
\begin{equation}
\begin{aligned}
	\delta(d\subs{lon}, d\subs{lat})  &= \arctan\left(\frac{d\subs{lon}}{R + d\subs{lat}}\right).
\end{aligned}
\end{equation}
The steered wheels are assumed to be connected to a single steering wheel, whose angle is denoted $\delta\subs{SWA}$. To distinguish clearly in figures, tire steering angles are always given in radians, and $\delta\subs{SWA}$ in degrees. %\footnote{In some vehicles, multiple unsteered axles can occur, such as \emph{tag axles} in trucks or coaches. In these cases, the longitudinal location of the chassis' origin can move between the axles, depending on the weight distribution.}
The velocity and angular rate of a wheel with radius $r\subs{tire}$ mounted at a point $[d\subs{lon}, d\subs{lat}]\transp$ relative to the chassis' origin is given by
\begin{align}
	v(d\subs{lon}, d\subs{lat}) &= \|\dot\traj\| \, \kappa \sqrt{d\subs{lon}^2+(\kappa^{-1} + d\subs{lat})^2}\,\quad\text{and}\\[5pt]
	\dot\rho(d\subs{lon}, d\subs{lat}) &=  \frac{v(d\subs{lon}, d\subs{lat})}{r\subs{tire}}
\end{align}
respectively; for the practical evaluations with a four-wheel vehicle, we use the indices $\delta\subs{FL}$ for the front left tire steering angle, or $v\subs{RR}$ for the rear right wheel speed over ground. Finally, the heading $\psi$ relative to a north direction $\boldsymbol n$, and yaw rate $\dot\psi$ are given by
\begin{align}
	\psi &= \operatorname{atan2}(\det[\boldsymbol n, \dot\traj], \boldsymbol n\transp \dot\traj) \quad \text{and}\\[5pt]
	\dot\psi &= \kappa \, \|\dot\traj\| = (-1)^R \frac{ \det[\dot\traj,\ddot{\traj}] }{ \|\dot\traj\|^2}.
\end{align}

\begin{figure}%
%\begin{subfigure}{\columnwidth}
%\includegraphics{images/c2-errors-pos6}
%\end{subfigure}

\begin{subfigure}{0.49\columnwidth}
\raggedleft
\begin{overpic}[%grid
]{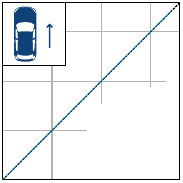}
\put(-35, -10){\parbox{1cm}{\raggedleft\footnotesize $0\,\unit{\frac{km}{h}}$}}
\put(7, -10){\parbox{1cm}{\raggedleft\footnotesize $50\,\unit{\frac{km}{h}}$}}
\put(35, -10){\parbox{1cm}{\raggedleft\footnotesize $100\,\unit{\frac{km}{h}}$}}
\put(65, -10){\parbox{1cm}{\raggedleft\footnotesize $150\,\unit{\frac{km}{h}}$}}
\put(-35, 80){\parbox{1cm}{\raggedleft\footnotesize $150\,\unit{\frac{km}{h}}$}}
\put(-35, 52){\parbox{1cm}{\raggedleft\footnotesize $100\,\unit{\frac{km}{h}}$}}
\put(-35, 25){\parbox{1cm}{\raggedleft\footnotesize $50\,\unit{\frac{km}{h}}$}}
\put(52, 15){\parbox{2cm}{
\scriptsize
$\mu = +0.68\,\unit{\tfrac{km}{h}}$\newline
$\sigma = 0.001\,\unit{\tfrac{km}{h}}$\newline
$m = 1.00$\newline
}}
\end{overpic}\\[10pt]
\caption{Estimated over simulated $v\subs{lon}$}%
\end{subfigure}
\hfill
\begin{subfigure}{0.49\columnwidth}
\raggedleft
\begin{overpic}[%grid
]{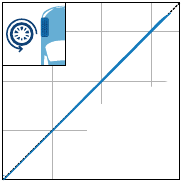}
\put(-35, -10){\parbox{1cm}{\raggedleft\footnotesize $0\,\unit{\frac{km}{h}}$}}
\put(7, -10){\parbox{1cm}{\raggedleft\footnotesize $50\,\unit{\frac{km}{h}}$}}
\put(35, -10){\parbox{1cm}{\raggedleft\footnotesize $100\,\unit{\frac{km}{h}}$}}
\put(65, -10){\parbox{1cm}{\raggedleft\footnotesize $150\,\unit{\frac{km}{h}}$}}
\put(-35, 80){\parbox{1cm}{\raggedleft\footnotesize $150\,\unit{\frac{km}{h}}$}}
\put(-35, 52){\parbox{1cm}{\raggedleft\footnotesize $100\,\unit{\frac{km}{h}}$}}
\put(-35, 25){\parbox{1cm}{\raggedleft\footnotesize $50\,\unit{\frac{km}{h}}$}}

\put(52, 15){\parbox{2cm}{
\scriptsize
$\mu = 0.81\,\unit{\frac{km}{h}}$\newline
$\sigma = 1.41\,\unit{\frac{km}{h}}$\newline
$m = 1.00$\newline
}}
\end{overpic}\\[10pt]
\caption{Estimated over simulated $v\subs{FL}$}%
\end{subfigure}\\[5pt]

\begin{subfigure}{0.49\columnwidth}
\raggedleft
\begin{overpic}[%grid
]{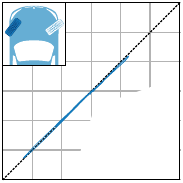}
\put(-9, -10){\parbox{1cm}{\raggedleft\footnotesize $-0.2$}}
\put(22, -10){\parbox{1cm}{\raggedleft\footnotesize $0$}}
\put(56, -10){\parbox{1cm}{\raggedleft\footnotesize $+0.2$}}
\put(-35, 15){\parbox{1cm}{\raggedleft\footnotesize $-0.2$}}
\put(-35, 31){\parbox{1cm}{\raggedleft\footnotesize $-0.1$}}
\put(-35, 46){\parbox{1cm}{\raggedleft\footnotesize $0$}}
\put(-35, 64){\parbox{1cm}{\raggedleft\footnotesize $+0.1$}}
\put(-35, 82){\parbox{1cm}{\raggedleft\footnotesize $+0.2$}}
\put(52, 15){\parbox{2cm}{
\scriptsize
$\mu = -0.002$\newline
$\sigma =0.003$\newline
$m = 0.93$\newline
}}
\end{overpic}\\[10pt]
\caption{Estimated over simulated $\delta\subs{FL}$}%
\end{subfigure}
\hfill
\begin{subfigure}{0.49\columnwidth}
\raggedleft
\begin{overpic}[%grid
]{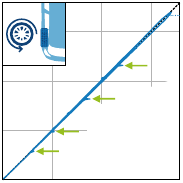}
\put(-35, -10){\parbox{1cm}{\raggedleft\footnotesize $0\,\unit{\frac{km}{h}}$}}
\put(7, -10){\parbox{1cm}{\raggedleft\footnotesize $50\,\unit{\frac{km}{h}}$}}
\put(35, -10){\parbox{1cm}{\raggedleft\footnotesize $100\,\unit{\frac{km}{h}}$}}
\put(65, -10){\parbox{1cm}{\raggedleft\footnotesize $150\,\unit{\frac{km}{h}}$}}
\put(-35, 80){\parbox{1cm}{\raggedleft\footnotesize $150\,\unit{\frac{km}{h}}$}}
\put(-35, 52){\parbox{1cm}{\raggedleft\footnotesize $100\,\unit{\frac{km}{h}}$}}
\put(-35, 25){\parbox{1cm}{\raggedleft\footnotesize $50\,\unit{\frac{km}{h}}$}}
\put(52, 15){\parbox{2cm}{
\scriptsize
$\mu = +1.14\,\unit{\tfrac{km}{h}}$\newline
$\sigma = 2.50\,\unit{\tfrac{km}{h}}$\newline
$m = 1.00$\newline
}}
\end{overpic}\\[10pt]
\caption{Estimated over simulated $v\subs{RL}$}%
\end{subfigure}%\\[5pt]

\caption{Selected simulation results from \cite{icmc} of a BMW 118i driving on the Hockenheimring race track, simulated in IPG CarMaker. With an overall adequate accuracy between estimated and simulated parameters, notable deviations include longitudinal wheel slip in the rear wheels (arrows in (d)) during sharp accelerations of the rear-wheel drive vehicle model.}%
\label{fig:carmaker-correlation}%
%\vspace{-10pt}
\end{figure}
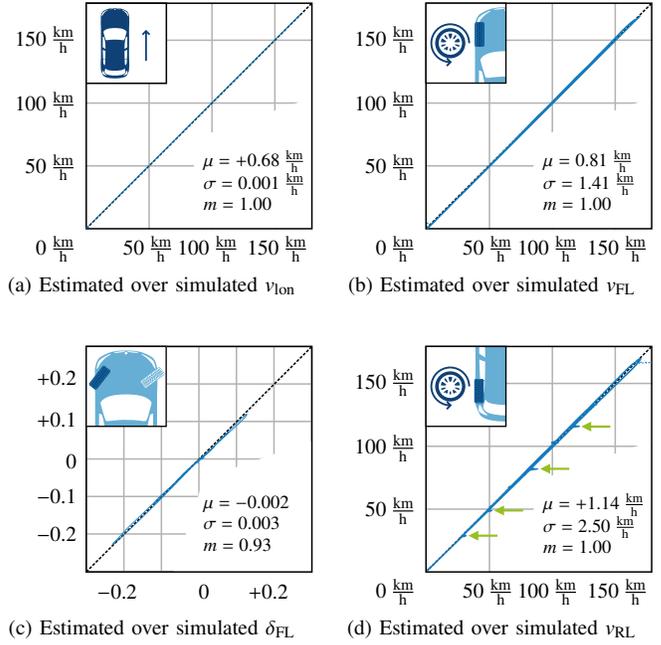

%\begin{equation}
%\left.\begin{matrix}a\subs{lat} = \kappa \, v^2\\[5pt]
%a\subs{lon} = \dee v / \dee t
%\end{matrix}\quad\right\}\quad
%\boldsymbol a = a\subs{lon}\, \boldsymbol T + a\subs{lat} \, \boldsymbol N
%\label{eq:}
%\end{equation}

\subsection{Forward Model}\label{sec:forward-model}

For various applications, it may be useful to solve the model for a trajectory, given control parameters such as $\delta\subs{SWA}$ and $v\subs{lon}$, via a differential equation. Due to the constraints of the model, any two independent parameters over time determine the system, e.g. two wheel speeds, but any third independent parameter overdetermines it. For example, for a vehicle with known geometric parameters and a steering function $f(\delta\subs{SWA})$ for a (virtual) center wheel mounted at a longitudinal offset of $\ell$ in front of the rear unsteered axle, we can solve for $\traj$ using $\delta\subs{SWA}$, $v\subs{lon}$ and the the local descriptions
\begin{equation}
\delta(\ell,0) = f(\delta\subs{SWA})\quad\text{and}\quad
\kappa = \tan\left(\frac{\delta(\ell,0)}{\ell}\right)
\end{equation}
to establish the set of differential equations
\begin{equation}
\frac{\dee \boldsymbol T }{ \dee t } =  \kappa \, v\subs{lon} \, \boldsymbol N \quad\text{and}\quad
\frac{\dee \boldsymbol \traj }{ \dee t } = v\subs{lon} \, \boldsymbol T
\end{equation}
where numerical applications must assure that $\smash{\|\boldsymbol T\| \equiv 1}$. Note that in both this variant of the forward model and in the analytic model, the wheel angles and speeds are inferred variables, which in this case may be extracted after solving for the trajectory, but are not required as state variables.

\section{SIMULATION RESULTS}\label{sec:simulation}

Simulation tests were conducted using IPG CarMaker in \cite{icmc} using a BMW 118i model on the Hockenheimring race track with speeds of up to $170\,\unit{km/h}$. The results show a very good correspondence between the simulated parameters, recorded at $1\,\unit{kHz}$, and the estimated results by the $C^2$ model, which was applied to the simulated trajectory.

Selected relevant results are reproduced in Fig.~\ref{fig:carmaker-correlation} as correlations between the $C^2$-estimated values (always on the vertical axis) over the simulation parameters (on the horizontal axis). Each plot gives three metrics for the accuracy of an estimation (which are also used in the real-world test drives): The mean error $\mu$ (which is ideally $\mu=0$, and negative for systematically underestimated parameters), the standard deviation $\sigma$ (unsigned, ideally $\sigma=0$) and the slope $m$ of the adjusted line which minimizes the equation
\begin{equation}
\int_{-\infty}^\infty \!\! \dee t\; \left|\;\;\chi\sups{sim}(t) \cdot m - \chi^{C^2}(t)\;\;\right|^2 \underset{m}{\longrightarrow} \min
\end{equation}
for any estimated parameter $\chi$, where values  $m \neq 1$ indicate that the parameter's scale is systematically biased.

In the simulation results, speeds match almost exactly with the $C^2$ model estimates. Front wheel speed accuracies match similarly well, while rear wheel speed accuracies show notable, isolated deviations (indicated by arrows in Fig.~\ref{fig:carmaker-correlation}d): Since the simulated BMW 118i is a rear-wheel drive, loss of traction at strong accelerations occurs at the rear wheels; if the vehicle's body is unable to keep up with the wheel rotations, the assumption of zero wheel slip is violated and speed is systematically underestimated by the $C^2$ model.

\section{REAL DRIVING TESTS}\label{sec:realtests}

\begin{figure}%
\begin{subfigure}{\columnwidth}
\includegraphics[width=\columnwidth]{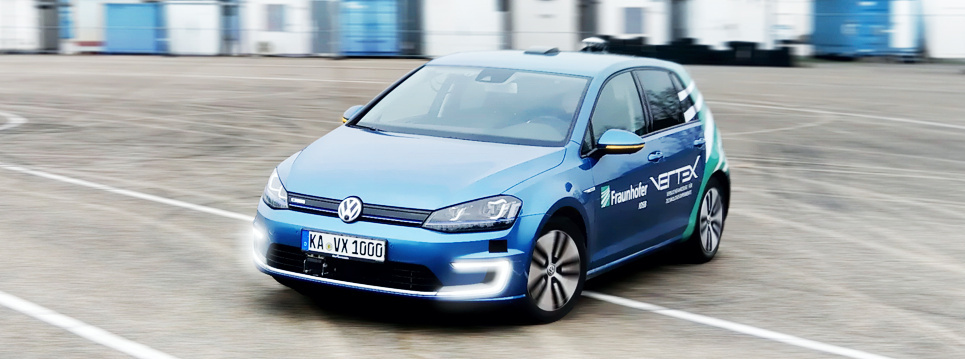}
\caption{The VW e-Golf VII test vehicle on the closed-off test track}%
\end{subfigure}\\[5pt]

\begin{subfigure}{\columnwidth}
\includegraphics{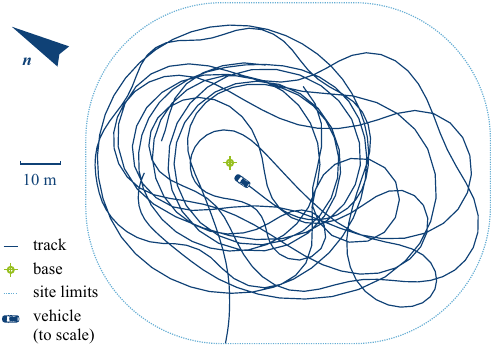}%
\caption{RTK GPS track of the vehicle}%
\end{subfigure}
\caption{Evaluations on the closed-off test track, Sec.~\ref{sec:tests-campusost}.}%
\label{fig:campusost-gps}%
\vspace{-10pt}
\end{figure}

To test the model under real driving conditions, tests were conducted on a closed-off test track (Sec.~\ref{sec:tests-campusost}) specifically for high dynamics, as well as on public roads (Sec.~\ref{sec:tests-interchange}) for more typical road conditions. The test vehicle is an electrical VW e-Golf VII equipped for automated driving \cite{vertex}, including access to CAN bus data (at a rate of about $50\,\unit{Hz}$) and RTK GPS with centimeter accuracy (at a rate of about $10\,\unit{Hz}$).

In each case, we apply the $C^2$ model to the RTK GPS trajectories to estimate the parameters steering wheel angle $\delta\subs{SWA}$, speed $v\subs{lon}$ and wheel speeds $v\subs{FL}$, $v\subs{FR}$, $v\subs{RL}$, $v\subs{RR}$, which are compared to the respective on-board vehicle sensors read from the CAN bus (Sec.~\ref{sec:tests-campusost-dynamic}). No on-board sensors are available for the tire steering angles $\delta\subs{FL}$ and $\delta\subs{FR}$; therefore, to evaluate these, we use the $C^2$ model estimates of $\delta\subs{FL}$ and $\delta\subs{FR}$, along with the corresponding recordings of $\delta\subs{SWA}$ from the CAN bus to estimate the steering function $\delta\subs{SWA}(\delta\subs{FL})$, and compare these to manual measurements of the wheels at rest (Sec.~\ref{sec:tests-campusost-swa}).

\subsection{Closed-off Test Track}\label{sec:tests-campusost}

Test drives under conditions that are unsuitable for public roads were performed on a closed-off site of the Test Area Autonomous Driving Baden-Württemberg (TAF-BW) located at KIT Campus Ost, Rintheimer Querallee 2, Karlsruhe, Germany.

\begin{figure}%
\raggedleft
\begin{subfigure}{0.85\columnwidth}
\raggedleft
\begin{overpic}[%grid
]{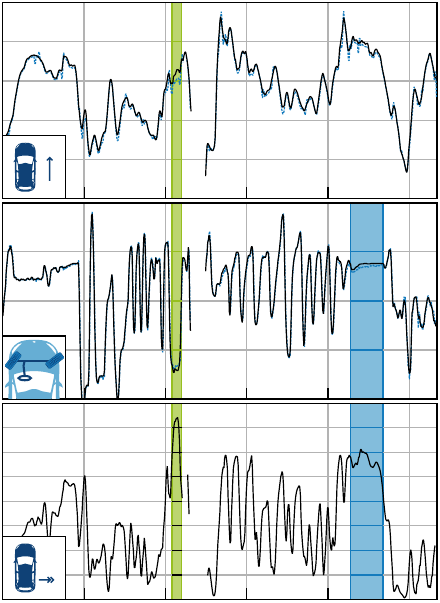}
\put(12, -3){\footnotesize $60\,\unit{s}$}
\put(25, -3){\footnotesize $120\,\unit{s}$}
\put(38, -3){\footnotesize $180\,\unit{s}$}
\put(52, -3){\footnotesize $240\,\unit{s}$}
\put(66, -3){\footnotesize $300\,\unit{s}$}
\put(-11, 99){\parbox{1cm}{\raggedleft\footnotesize $50\,\unit{\tfrac{km}{h}}$}}
\put(-11, 92){\parbox{1cm}{\raggedleft\footnotesize $40\,\unit{\tfrac{km}{h}}$}}
\put(-11, 85){\parbox{1cm}{\raggedleft\footnotesize $30\,\unit{\tfrac{km}{h}}$}}
\put(-12, 84){\small $v\subs{lon}$}
\put(-11, 79){\parbox{1cm}{\raggedleft\footnotesize $20\,\unit{\tfrac{km}{h}}$}}
\put(-11, 72.5){\parbox{1cm}{\raggedleft\footnotesize $10\,\unit{\tfrac{km}{h}}$}}
\put(-11, 65){\parbox{1cm}{\raggedleft\footnotesize $360^\circ$}}
\put(-11, 57){\parbox{1cm}{\raggedleft\footnotesize $180^\circ$}}
\put(-11, 49){\parbox{1cm}{\raggedleft\footnotesize $0^\circ$}}
\put(-12, 49){\small $\delta\subs{SWA}$}
\put(-11, 41){\parbox{1cm}{\raggedleft\footnotesize $-180^\circ$}}
\put(-11, 33){\parbox{1cm}{\raggedleft\footnotesize $-360^\circ$}}
\put(-11, 24){\parbox{1cm}{\raggedleft\footnotesize $6\,\unit{\tfrac{m}{s^2}}$}}
\put(-11, 16){\parbox{1cm}{\raggedleft\footnotesize $4\,\unit{\tfrac{m}{s^2}}$}}
\put(-12, 16){\small $a\subs{lat}$}
\put(-11, 7.5){\parbox{1cm}{\raggedleft\footnotesize $2\,\unit{\tfrac{m}{s^2}}$}}
\put(-11, -1){\parbox{1cm}{\raggedleft\footnotesize $0\,\unit{\tfrac{m}{s^2}}$}}
\end{overpic}\\[8pt]
\caption{Complete data sequences of the test drive}
\label{fig:campusost-time-all}
\end{subfigure}\\[5pt]

\begin{subfigure}{0.85\columnwidth}
\raggedleft
\begin{overpic}[%grid
]{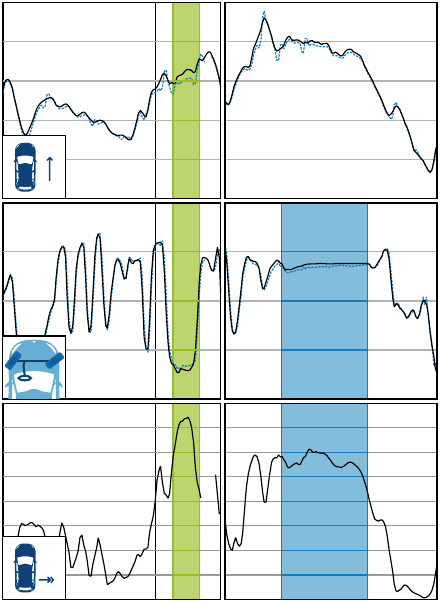}
\put(0, -3){\footnotesize $80\,\unit{s}$}
\put(30, -3){\footnotesize $140\,\unit{s}$}
\put(37, -3){\footnotesize $240\,\unit{s}$}
\put(66, -3){\footnotesize $300\,\unit{s}$}
\put(-11, 99){\parbox{1cm}{\raggedleft\footnotesize $50\,\unit{\tfrac{km}{h}}$}}
\put(-11, 92){\parbox{1cm}{\raggedleft\footnotesize $40\,\unit{\tfrac{km}{h}}$}}
\put(-11, 85){\parbox{1cm}{\raggedleft\footnotesize $30\,\unit{\tfrac{km}{h}}$}}
\put(-12, 84){\small $v\subs{lon}$}
\put(-11, 79){\parbox{1cm}{\raggedleft\footnotesize $20\,\unit{\tfrac{km}{h}}$}}
\put(-11, 72.5){\parbox{1cm}{\raggedleft\footnotesize $10\,\unit{\tfrac{km}{h}}$}}
\put(-11, 65){\parbox{1cm}{\raggedleft\footnotesize $360^\circ$}}
\put(-11, 57){\parbox{1cm}{\raggedleft\footnotesize $180^\circ$}}
\put(-11, 49){\parbox{1cm}{\raggedleft\footnotesize $0^\circ$}}
\put(-12, 49){\small $\delta\subs{SWA}$}
\put(-11, 41){\parbox{1cm}{\raggedleft\footnotesize $-180^\circ$}}
\put(-11, 33){\parbox{1cm}{\raggedleft\footnotesize $-360^\circ$}}
\put(-11, 24){\parbox{1cm}{\raggedleft\footnotesize $6\,\unit{\tfrac{m}{s^2}}$}}
\put(-11, 16){\parbox{1cm}{\raggedleft\footnotesize $4\,\unit{\tfrac{m}{s^2}}$}}
\put(-12, 16){\small $a\subs{lat}$}
\put(-11, 7.5){\parbox{1cm}{\raggedleft\footnotesize $2\,\unit{\tfrac{m}{s^2}}$}}
\put(-11, -1){\parbox{1cm}{\raggedleft\footnotesize $0\,\unit{\tfrac{m}{s^2}}$}}
\end{overpic}\\[6pt]
\caption{Details near shaded regions of strong errors}
\label{fig:campusost-time-detail}
\end{subfigure}\\

\caption{Overview of measured parameters (solid black) vs. $C^2$ model estimates (dashed blue) over time (a) and extracts around the two regions of strong errors (shaded) in $\delta\subs{SWA}^{C^2}$ and $v\subs{lon}^{C^2}$, which correspond with high lateral accelerations.}%
\label{fig:campusost-time}%
\end{figure}
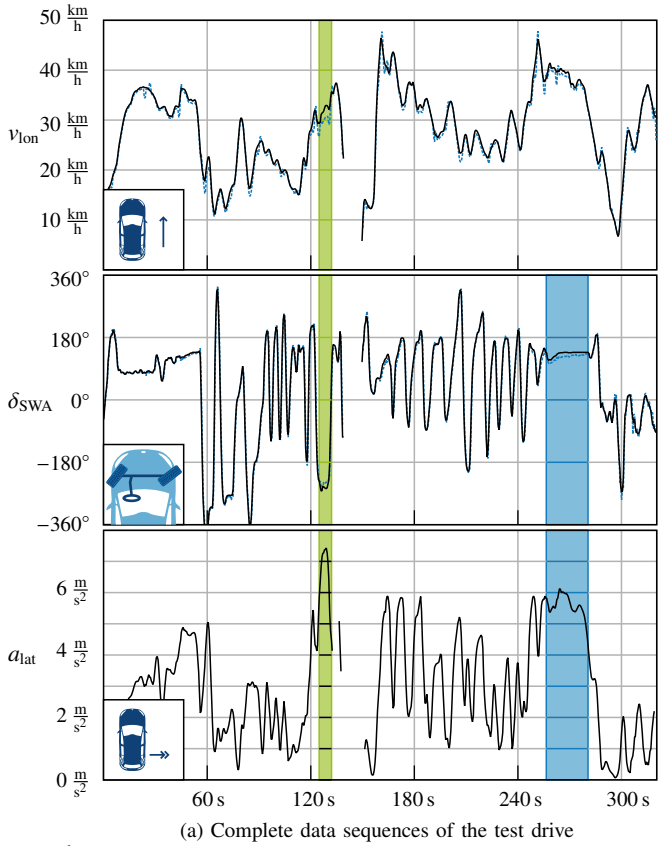
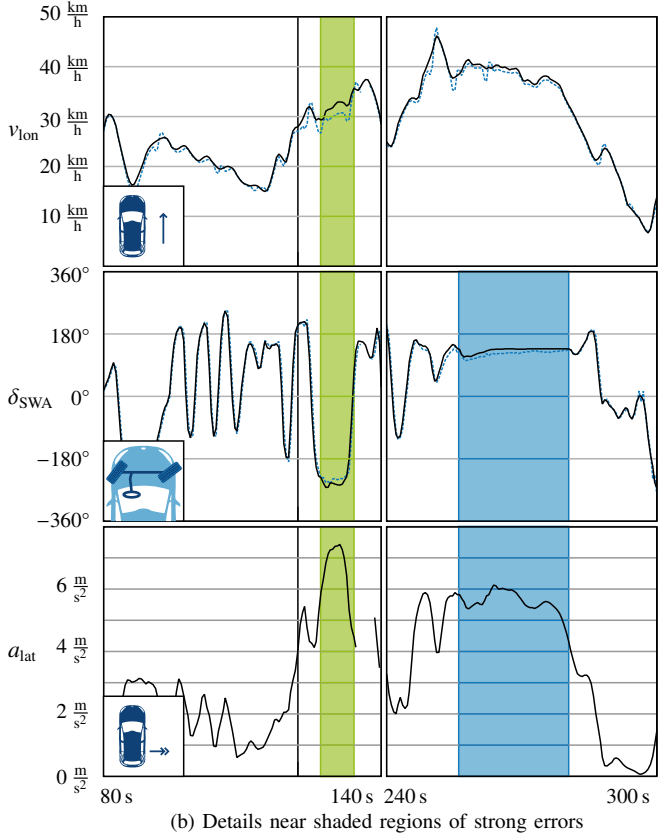

\subsubsection{Comparison with On-Board Sensors}\label{sec:tests-campusost-dynamic}

A total of about $300\,\unit{s}$ of data was collected during test drives, with longitudinal accelerations of $a\subs{lon}$ within $[-3.6\,\unit{m/s^2}, 3.2\,\unit{m/s^2}]$ and lateral accelerations $|a\subs{lat}|$ of above $7.5\,\unit{m/s^2}$, with a maximum speed of $48\,\unit{km/h}$. The RTK GPS track is given in Fig.~\ref{fig:campusost-gps}, and contains steering wheel angles of $|\delta\subs{SWA}| > 360^\circ$.

A direct comparison of CAN measurements and $C^2$ model estimates is given in Fig.~\ref{fig:campusost-time}: It can be seen that despite the unusually strong dynamics of the trajectory, the steering wheel angle $\delta\subs{SWA}$ is estimated with good accuracy most of the time. Two significant exceptions are highlighted in the figure: In both cases, the absolute angle is underestimated by more than $5\,\%$, while the typical error is considerably smaller. Both cases (shown in detail in Fig.~\ref{fig:campusost-time-detail}) occur during times of strong lateral accelerations. This indicates that the (front-wheel drive) vehicle is understeering, violating the assumption of zero lateral wheel velocity of the $C^2$ model.

\begin{figure}[t]%
%\begin{subfigure}{\columnwidth}
%\includegraphics{images/c2-errors-pos6}
%\end{subfigure}

\begin{subfigure}{0.49\columnwidth}
\raggedleft
\begin{overpic}[%grid
]{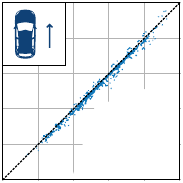}
\put(-35, -10){\parbox{1cm}{\raggedleft\footnotesize $0\,\unit{\tfrac{km}{h}}$}}
\put(20, -10){\parbox{1cm}{\raggedleft\footnotesize $20\,\unit{\tfrac{km}{h}}$}}
\put(60, -10){\parbox{1cm}{\raggedleft\footnotesize $40\,\unit{\tfrac{km}{h}}$}}
\put(-35, 18){\parbox{1cm}{\raggedleft\footnotesize $10\,\unit{\tfrac{km}{h}}$}}
\put(-35, 37){\parbox{1cm}{\raggedleft\footnotesize $20\,\unit{\tfrac{km}{h}}$}}
\put(-35, 56){\parbox{1cm}{\raggedleft\footnotesize $30\,\unit{\tfrac{km}{h}}$}}
\put(-35, 77){\parbox{1cm}{\raggedleft\footnotesize $40\,\unit{\tfrac{km}{h}}$}}
\put(52, 15){\parbox{2cm}{
\scriptsize
$\mu = +0.54\,\unit{\tfrac{km}{h}}$\newline
$\sigma = 0.78\,\unit{\tfrac{km}{h}}$\newline
$m = 1.00$\newline
}}
\end{overpic}\\[10pt]
\caption{Estimated over measured $v\subs{lon}$}%
\end{subfigure}
\hfill
\begin{subfigure}{0.49\columnwidth}
\raggedleft
\begin{overpic}[%grid
]{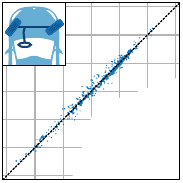}
\put(-3, -10){\parbox{1cm}{\raggedleft\footnotesize $-360^\circ$}}
\put(25, -10){\parbox{1cm}{\raggedleft\footnotesize $0^\circ$}}
\put(57, -10){\parbox{1cm}{\raggedleft\footnotesize $+360^\circ$}}
\put(-35, 80){\parbox{1cm}{\raggedleft\footnotesize $+360^\circ$}}
\put(-35, 63){\parbox{1cm}{\raggedleft\footnotesize $+180^\circ$}}
\put(-35, 46){\parbox{1cm}{\raggedleft\footnotesize $0^\circ$}}
\put(-35, 31){\parbox{1cm}{\raggedleft\footnotesize $-180^\circ$}}
\put(-35, 15){\parbox{1cm}{\raggedleft\footnotesize $-360^\circ$}}

\put(52, 15){\parbox{2cm}{
\scriptsize
$\mu = +1.1^\circ$\newline
$\sigma = 16.7^\circ$\newline
$m = 0.99$\newline
}}
\end{overpic}\\[10pt]
\caption{Estimated over measured $\delta\subs{SWA}$}%
\end{subfigure}\\%[3pt]

\begin{subfigure}{0.49\columnwidth}
\raggedleft
\begin{overpic}[%grid
]{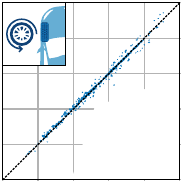}
\put(-35, -10){\parbox{1cm}{\raggedleft\footnotesize $0\,\unit{\tfrac{km}{h}}$}}
\put(20, -10){\parbox{1cm}{\raggedleft\footnotesize $20\,\unit{\tfrac{km}{h}}$}}
\put(60, -10){\parbox{1cm}{\raggedleft\footnotesize $40\,\unit{\tfrac{km}{h}}$}}
\put(-35, 18){\parbox{1cm}{\raggedleft\footnotesize $10\,\unit{\tfrac{km}{h}}$}}
\put(-35, 37){\parbox{1cm}{\raggedleft\footnotesize $20\,\unit{\tfrac{km}{h}}$}}
\put(-35, 56){\parbox{1cm}{\raggedleft\footnotesize $30\,\unit{\tfrac{km}{h}}$}}
\put(-35, 77){\parbox{1cm}{\raggedleft\footnotesize $40\,\unit{\tfrac{km}{h}}$}}
\put(52, 15){\parbox{2cm}{
\scriptsize
$\mu = -0.02\,\unit{\tfrac{km}{h}}$\newline
$\sigma =0.69\,\unit{\tfrac{km}{h}}$\newline
$m = 0.98$\newline
}}
\end{overpic}\\[10pt]
\caption{Estimated over measured $v\subs{FL}$}%
\end{subfigure}
\hfill
\begin{subfigure}{0.49\columnwidth}
\raggedleft
\begin{overpic}[%grid
]{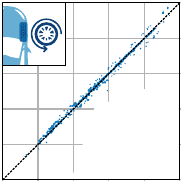}
\put(-35, -10){\parbox{1cm}{\raggedleft\footnotesize $0\,\unit{\tfrac{km}{h}}$}}
\put(20, -10){\parbox{1cm}{\raggedleft\footnotesize $20\,\unit{\tfrac{km}{h}}$}}
\put(60, -10){\parbox{1cm}{\raggedleft\footnotesize $40\,\unit{\tfrac{km}{h}}$}}
\put(-35, 18){\parbox{1cm}{\raggedleft\footnotesize $10\,\unit{\tfrac{km}{h}}$}}
\put(-35, 37){\parbox{1cm}{\raggedleft\footnotesize $20\,\unit{\tfrac{km}{h}}$}}
\put(-35, 56){\parbox{1cm}{\raggedleft\footnotesize $30\,\unit{\tfrac{km}{h}}$}}
\put(-35, 77){\parbox{1cm}{\raggedleft\footnotesize $40\,\unit{\tfrac{km}{h}}$}}
\put(52, 15){\parbox{2cm}{
\scriptsize
$\mu = -0.03\,\unit{\tfrac{km}{h}}$\newline
$\sigma = 0.71\,\unit{\tfrac{km}{h}}$\newline
$m = 0.98$\newline
}}
\end{overpic}\\[10pt]
\caption{Estimated over measured $v\subs{FR}$}%
\end{subfigure}\\%[3pt]

\begin{subfigure}{0.49\columnwidth}
\raggedleft
\begin{overpic}[%grid
]{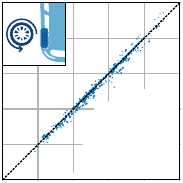}
\put(-35, -10){\parbox{1cm}{\raggedleft\footnotesize $0\,\unit{\tfrac{km}{h}}$}}
\put(20, -10){\parbox{1cm}{\raggedleft\footnotesize $20\,\unit{\tfrac{km}{h}}$}}
\put(60, -10){\parbox{1cm}{\raggedleft\footnotesize $40\,\unit{\tfrac{km}{h}}$}}
\put(-35, 18){\parbox{1cm}{\raggedleft\footnotesize $10\,\unit{\tfrac{km}{h}}$}}
\put(-35, 37){\parbox{1cm}{\raggedleft\footnotesize $20\,\unit{\tfrac{km}{h}}$}}
\put(-35, 56){\parbox{1cm}{\raggedleft\footnotesize $30\,\unit{\tfrac{km}{h}}$}}
\put(-35, 77){\parbox{1cm}{\raggedleft\footnotesize $40\,\unit{\tfrac{km}{h}}$}}
\put(52, 15){\parbox{2cm}{
\scriptsize
$\mu = +0.07\,\unit{\tfrac{km}{h}}$\newline
$\sigma = 1.86\,\unit{\tfrac{km}{h}}$\newline
$m = 1.01$\newline
}}
\end{overpic}\\[10pt]
\caption{Estimated over measured $v\subs{RL}$}%
\end{subfigure}
\hfill
\begin{subfigure}{0.49\columnwidth}
\raggedleft
\begin{overpic}[%grid
]{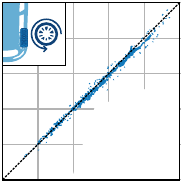}
\put(-35, -10){\parbox{1cm}{\raggedleft\footnotesize $0\,\unit{\tfrac{km}{h}}$}}
\put(20, -10){\parbox{1cm}{\raggedleft\footnotesize $20\,\unit{\tfrac{km}{h}}$}}
\put(60, -10){\parbox{1cm}{\raggedleft\footnotesize $40\,\unit{\tfrac{km}{h}}$}}
\put(-35, 18){\parbox{1cm}{\raggedleft\footnotesize $10\,\unit{\tfrac{km}{h}}$}}
\put(-35, 37){\parbox{1cm}{\raggedleft\footnotesize $20\,\unit{\tfrac{km}{h}}$}}
\put(-35, 56){\parbox{1cm}{\raggedleft\footnotesize $30\,\unit{\tfrac{km}{h}}$}}
\put(-35, 77){\parbox{1cm}{\raggedleft\footnotesize $40\,\unit{\tfrac{km}{h}}$}}
\put(52, 15){\parbox{2cm}{
\scriptsize
$\mu = +0.53\,\unit{\tfrac{km}{h}}$\newline
$\sigma = 0.87\,\unit{\tfrac{km}{h}}$\newline
$m = 0.96$\newline
}}
\end{overpic}\\[10pt]
\caption{Estimated over measured $v\subs{RR}$}%
\end{subfigure}%\\[3pt]

\caption{Accuracy of parameters at the closed-off test track data set, with parameter accuracies significantly reduced compared to the simulation results in Fig.~\ref{fig:carmaker-correlation}, yet within the expected range of position measurement uncertainties.}%
\label{fig:campusost-correlation}%
\vspace{-10pt}
\end{figure}

Overall accuracies are given in Fig.~\ref{fig:campusost-correlation}, as in Fig.~\ref{fig:carmaker-correlation}, as correlations between the $C^2$-estimated values (always on the vertical axis) over the CAN measurements (always on the horizontal axis); parameters $\mu$, $\sigma$, $m$ are as introduced in Sec.~\ref{sec:simulation}. Comparison with results from \cite{icmc} shows a notably reduced accuracy for any measurement; yet, the overall accuracy scales consistently with the degradation of quality in speed $v\subs{lon}$ (the most trivial parameter, estimated almost perfectly in the simulation results), indicating that the reduction in accuracy is dominated by measurement uncertainties, as opposed to model limitations.

\subsubsection{Evaluation of Tire Steering Angles}\label{sec:tests-campusost-swa}

With no sensors to measure tire steering angles with the vehicle in motion, the tire angle estimates were evaluated by establishing the estimated steering function $\delta\subs{SWA}(\delta\subs{FL})$ (or $-\delta\subs{SWA}(-\delta\subs{FR})$) from the test track recordings, and comparing the result to a manual measurement on a turntable with the vehicle at rest. For the former, the $C^2$ estimates $\delta\subs{FL}^{C^2}, \delta\subs{FR}^{C^2}$  during the driving tests are correlated with the CAN bus recordings of the steering wheel angle $\delta\subs{SWA}\sups{CAN}$. A numeric adjustment of cubic polynomial parameters yields the \emph{estimated} steering function.

\begin{figure}%
\begin{subfigure}{30mm}
\includegraphics{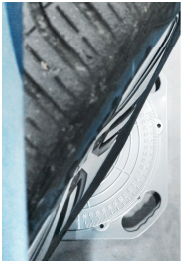}
\caption{Wheel angle measurements by turntable}
\end{subfigure}
\hfill
\begin{subfigure}{46mm}
\raggedleft
\begin{overpic}[%grid
]{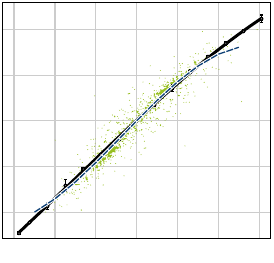}%
%\put(9, 3){\footnotesize $-0.4$}
\put(13, 3){\footnotesize $-0.4$}
\put(28, 3){\footnotesize $-0.2$}
\put(49, 3){\footnotesize $0$}
\put(58, 3){\footnotesize $+0.2$}
\put(73, 3){\footnotesize $+0.4$}
%\put(82, 3){\footnotesize $+0.4$}
%
%\put(-22, 5){\parbox{1cm}{\raggedleft\footnotesize $-600^\circ$}}
\put(-22, 17){\parbox{1cm}{\raggedleft\footnotesize $-400^\circ$}}
\put(-22, 33){\parbox{1cm}{\raggedleft\footnotesize $-200^\circ$}}
\put(-22, 50){\parbox{1cm}{\raggedleft\footnotesize $0^\circ$}}
\put(-22, 67){\parbox{1cm}{\raggedleft\footnotesize $+200^\circ$}}
\put(-22, 83){\parbox{1cm}{\raggedleft\footnotesize $+400^\circ$}}
%\put(-22, 80){\parbox{1cm}{\raggedleft\footnotesize $+600^\circ$}}
\end{overpic}
\caption{Combined angle measurements for $\delta\subs{SWA}(\delta\subs{FL})$ and $-\delta\subs{SWA}(-\delta\subs{FR})$}
\end{subfigure}
\caption{The relation of steering wheel angles $\delta\subs{SWA}$ to wheel (i.e. tire) angles $\delta\subs{FL}, \delta\subs{FR}$ was evaluated by comparing recorded CAN signals of $\delta\subs{SWA}$ to manual turntable measurements (a) and $C^2$ estimates respectively. Figure (b) shows $\delta\subs{SWA}(\delta\subs{FL})$ and $-\delta\subs{SWA}(-\delta\subs{FR})$ from the $C^2$ estimates (point scatter plot) and their adjusted cubic polynomial (dashed line), along with manual measurements by a turntable (black circles) and their adjusted cubic polynomial (solid black line).}%
\label{fig:turntable}%
%\vspace{-10pt}
\end{figure}
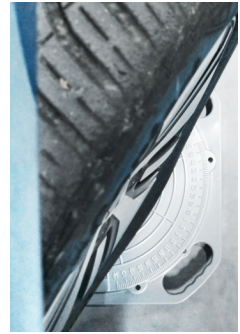
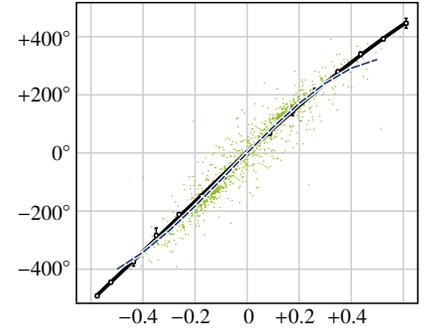

On the turntable, the tire steering angles are measured manually, along with the CAN measurements of $\delta\subs{SWA}\sups{CAN}$; again, an adjustment of cubic polynomial parameters yields the \emph{measured} steering function. The results are shown in Fig.~\ref{fig:turntable}. The slope of the polynomials, $m^{C^2}\! \approx 859^\circ$ vs. $m\sups{meas} \approx 830^\circ$ differs by around $5\,\%$, but the trend at extreme angles with few measurements deviates notably. Whether agreement for moderate $\delta\subs{SWA}$ holds for a moving vehicle cannot be concluded based on the data, and is left for future work.

\begin{figure}[t]%

\begin{subfigure}{\columnwidth}
\includegraphics{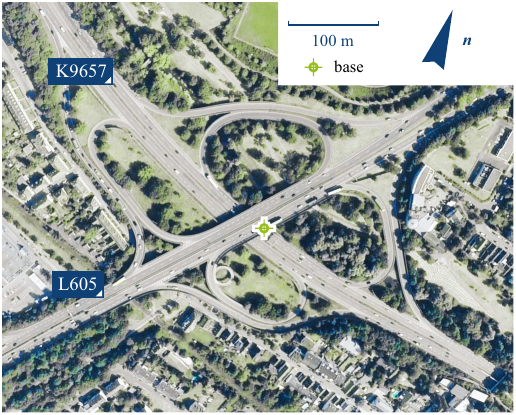}%
\caption{Aerial view of the interchange near Karlsruhe, Germany}%
\label{fig:interchange-aer}%
\end{subfigure}\\[10pt]

\begin{subfigure}{\columnwidth}
\includegraphics[width=\columnwidth]{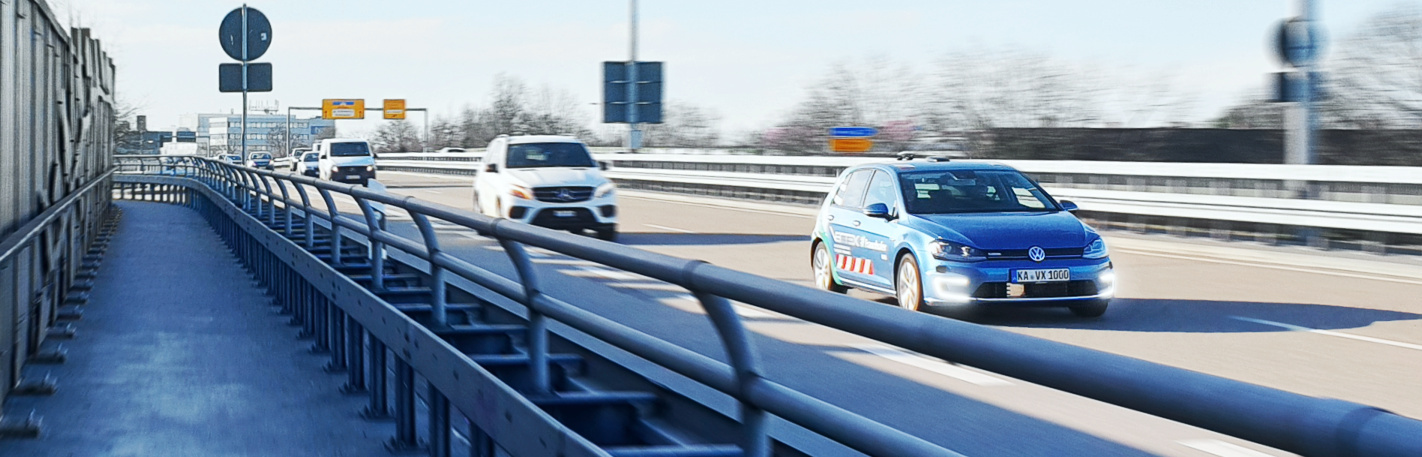}
\caption{Ground view from the position of the RTK GPS base, facing south}
\end{subfigure}
\caption{Views of the interchange test drive area used and discussed in Sec.~\ref{sec:tests-interchange}.}%
\vspace{-10pt}
\end{figure}

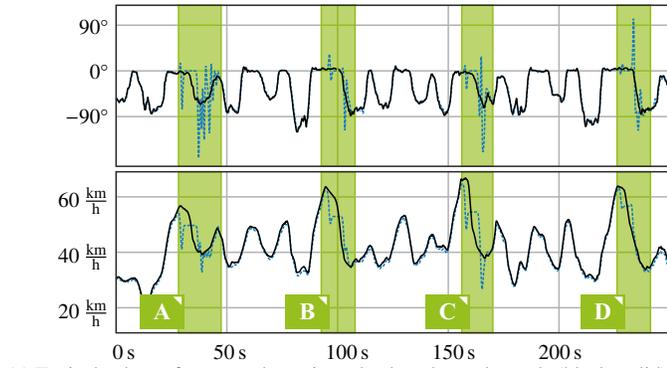
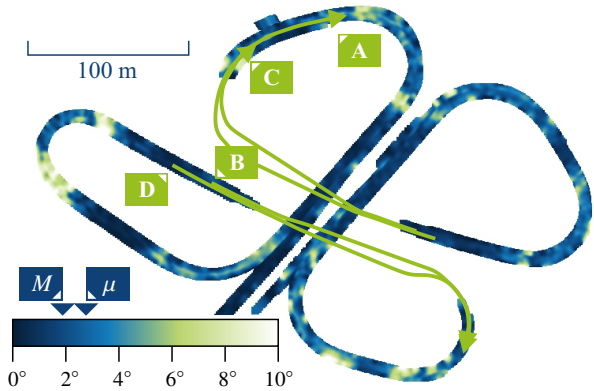
\begin{figure}%
%\begin{subfigure}{\columnwidth}
%\includegraphics{images/c2-errors-pos6}
%\end{subfigure}

\begin{subfigure}{\columnwidth}
\raggedleft
\begin{overpic}[%grid
]{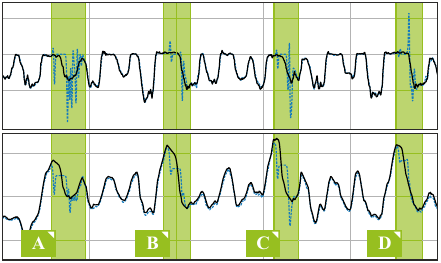}
\put(0, -4){\footnotesize $0\,\unit{s}$}
\put(18, -4){\footnotesize $50\,\unit{s}$}
\put(38, -4){\footnotesize $100\,\unit{s}$}
\put(57, -4){\footnotesize $150\,\unit{s}$}
\put(76, -4){\footnotesize $200\,\unit{s}$}
%\put(92, -4){\footnotesize $250\,\unit{s}$}
%
\put(-14, 3){\parbox{1cm}{\raggedleft\footnotesize $20\,\unit{\tfrac{km}{h}}$}}
\put(-14, 13){\parbox{1cm}{\raggedleft\footnotesize $40\,\unit{\tfrac{km}{h}}$}}
\put(-14, 23){\parbox{1cm}{\raggedleft\footnotesize $60\,\unit{\tfrac{km}{h}}$}}
\put(-14, 46){\parbox{1cm}{\raggedleft\footnotesize $0^\circ$}}
\put(-14, 38){\parbox{1cm}{\raggedleft\footnotesize $-90^\circ$}}
\put(-14, 54){\parbox{1cm}{\raggedleft\footnotesize $90^\circ$}}
\end{overpic}\\[5pt]
\caption{Typical values of measured steering wheel angles and speeds (black, solid), vs. $C^2$ model estimates (blue, dashed). Periodic regions of extreme deviations (labeled \textbf{A}--\textbf{D}) occur when the RTK GPS signal is occluded under the highway overpass (cf. (b)). These regions were excluded from further processing steps.}%
\label{fig:interchange-quality-delta-v}
\end{subfigure}\\%[15pt]

%\begin{subfigure}{\columnwidth}
%\raggedleft
%\begin{overpic}[%grid
%]{images/c2-gaps-plot-small-2}
%\end{overpic}\\[5pt]
%\caption{Errors from $\delta_{C^2}, v_{C^2}$ over $s$}%
%\label{fig:interchange-quality-delete}
%\end{subfigure}\\[10pt]
%

\begin{subfigure}{\columnwidth}
\raggedleft
\begin{overpic}[%grid
]{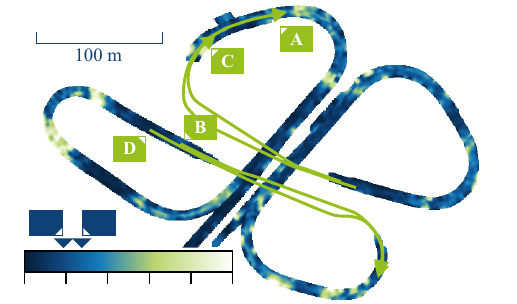}
\put(7.5, 14.5){\small \color{white} $M$}
\put(18.5, 14.5){\small \color{white} $\mu$}
\put(4, 0){\footnotesize $0^\circ$}
\put(12, 0){\footnotesize $2^\circ$}
\put(20, 0){\footnotesize $4^\circ$}
\put(28, 0){\footnotesize $6^\circ$}
\put(36, 0){\footnotesize $8^\circ$}
\put(43, 0){\footnotesize $10^\circ$}
\end{overpic}\\[5pt]
\caption{Removed trajectories \textbf{A}--\textbf{D} from (a), and average absolute errors of $\delta$ by positions after removing regions around extreme outliers, with overall median error $M$ and mean error $\mu$ indicated.}%
\label{fig:interchange-quality-delta-errors}
\end{subfigure}

\caption{Qualitative examples of results at the interchange (a), removal of outliers due to occluded RTK GPS (a--b) and local accuracy of steering wheel angle estimates (b).}%
\label{fig:interchange-quality}%

\vspace{-10pt}
\end{figure}

\begin{figure}[t!]%
%\begin{subfigure}{\columnwidth}
%\includegraphics{images/c2-errors-pos6}
%\end{subfigure}

\begin{subfigure}{0.49\columnwidth}
\raggedleft
\begin{overpic}[%grid
]{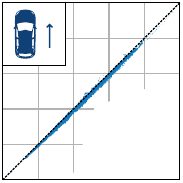}
\put(-35, -10){\parbox{1cm}{\raggedleft\footnotesize $0\,\unit{\tfrac{km}{h}}$}}
\put(20, -10){\parbox{1cm}{\raggedleft\footnotesize $40\,\unit{\tfrac{km}{h}}$}}
\put(60, -10){\parbox{1cm}{\raggedleft\footnotesize $80\,\unit{\tfrac{km}{h}}$}}
\put(-35, 18){\parbox{1cm}{\raggedleft\footnotesize $20\,\unit{\tfrac{km}{h}}$}}
\put(-35, 37){\parbox{1cm}{\raggedleft\footnotesize $40\,\unit{\tfrac{km}{h}}$}}
\put(-35, 56){\parbox{1cm}{\raggedleft\footnotesize $60\,\unit{\tfrac{km}{h}}$}}
\put(-35, 77){\parbox{1cm}{\raggedleft\footnotesize $80\,\unit{\tfrac{km}{h}}$}}
\put(52, 15){\parbox{2cm}{
\scriptsize
$\mu = +0.87\,\unit{\tfrac{km}{h}}$\newline
$\sigma = 1.03\,\unit{\tfrac{km}{h}}$\newline
$m = 0.98$\newline
}}
\end{overpic}\\[10pt]
\caption{Estimated over measured $v\subs{lon}$}%
\end{subfigure}
\hfill
\begin{subfigure}{0.49\columnwidth}
\raggedleft
\begin{overpic}[%grid
]{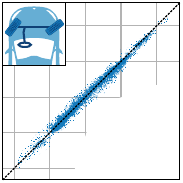}
\put(-15, -10){\parbox{1cm}{\raggedleft\footnotesize $-135^\circ$}}
\put(5, -10){\parbox{1cm}{\raggedleft\footnotesize $-90^\circ$}}
\put(25, -10){\parbox{1cm}{\raggedleft\footnotesize $-45^\circ$}}
\put(39, -10){\parbox{1cm}{\raggedleft\footnotesize $0^\circ$}}
\put(63, -10){\parbox{1cm}{\raggedleft\footnotesize $+45^\circ$}}
\put(-35, 83){\parbox{1cm}{\raggedleft\footnotesize $+45^\circ$}}
\put(-35, 63){\parbox{1cm}{\raggedleft\footnotesize $0^\circ$}}
\put(-35, 45){\parbox{1cm}{\raggedleft\footnotesize $-45^\circ$}}
\put(-35, 25){\parbox{1cm}{\raggedleft\footnotesize $-90^\circ$}}
\put(-35, 5){\parbox{1cm}{\raggedleft\footnotesize $-135^\circ$}}

\put(52, 15){\parbox{2cm}{
\scriptsize
$\mu = +0.5^\circ$\newline
$\sigma = 3.9^\circ$\newline
$m = 0.97$\newline
}}
\end{overpic}\\[10pt]
\caption{Estimated over measured $\delta\subs{SWA}$}%
\end{subfigure}\\[-2pt]

\begin{subfigure}{0.49\columnwidth}
\raggedleft
\begin{overpic}[%grid
]{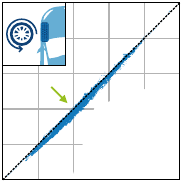}
\put(-35, -10){\parbox{1cm}{\raggedleft\footnotesize $0\,\unit{\tfrac{km}{h}}$}}
\put(20, -10){\parbox{1cm}{\raggedleft\footnotesize $40\,\unit{\tfrac{km}{h}}$}}
\put(60, -10){\parbox{1cm}{\raggedleft\footnotesize $80\,\unit{\tfrac{km}{h}}$}}
\put(-35, 18){\parbox{1cm}{\raggedleft\footnotesize $20\,\unit{\tfrac{km}{h}}$}}
\put(-35, 37){\parbox{1cm}{\raggedleft\footnotesize $40\,\unit{\tfrac{km}{h}}$}}
\put(-35, 56){\parbox{1cm}{\raggedleft\footnotesize $60\,\unit{\tfrac{km}{h}}$}}
\put(-35, 77){\parbox{1cm}{\raggedleft\footnotesize $80\,\unit{\tfrac{km}{h}}$}}
\put(52, 15){\parbox{2cm}{
\scriptsize
$\mu = +1.32\,\unit{\tfrac{km}{h}}$\newline
$\sigma = 1.63\,\unit{\tfrac{km}{h}}$\newline
$m = 1.01$\newline
}}
\end{overpic}\\[10pt]
\caption{Estimated over measured $v\subs{FL}$}%
\end{subfigure}
\hfill
\begin{subfigure}{0.49\columnwidth}
\raggedleft
\begin{overpic}[%grid
]{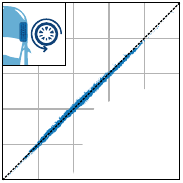}
\put(-35, -10){\parbox{1cm}{\raggedleft\footnotesize $0\,\unit{\tfrac{km}{h}}$}}
\put(20, -10){\parbox{1cm}{\raggedleft\footnotesize $40\,\unit{\tfrac{km}{h}}$}}
\put(60, -10){\parbox{1cm}{\raggedleft\footnotesize $80\,\unit{\tfrac{km}{h}}$}}
\put(-35, 18){\parbox{1cm}{\raggedleft\footnotesize $20\,\unit{\tfrac{km}{h}}$}}
\put(-35, 37){\parbox{1cm}{\raggedleft\footnotesize $40\,\unit{\tfrac{km}{h}}$}}
\put(-35, 56){\parbox{1cm}{\raggedleft\footnotesize $60\,\unit{\tfrac{km}{h}}$}}
\put(-35, 77){\parbox{1cm}{\raggedleft\footnotesize $80\,\unit{\tfrac{km}{h}}$}}
\put(52, 15){\parbox{2cm}{
\scriptsize
$\mu = -0.09\,\unit{\tfrac{km}{h}}$\newline
$\sigma = 0.92\,\unit{\tfrac{km}{h}}$\newline
$m = 0.96$\newline
}}
\end{overpic}\\[10pt]
\caption{Estimated over measured $v\subs{FR}$}%
\end{subfigure}\\[-2pt]

\begin{subfigure}{0.49\columnwidth}
\raggedleft
\begin{overpic}[%grid
]{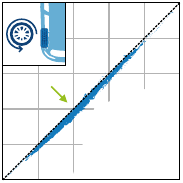}
\put(-35, -10){\parbox{1cm}{\raggedleft\footnotesize $0\,\unit{\tfrac{km}{h}}$}}
\put(20, -10){\parbox{1cm}{\raggedleft\footnotesize $40\,\unit{\tfrac{km}{h}}$}}
\put(60, -10){\parbox{1cm}{\raggedleft\footnotesize $80\,\unit{\tfrac{km}{h}}$}}
\put(-35, 18){\parbox{1cm}{\raggedleft\footnotesize $20\,\unit{\tfrac{km}{h}}$}}
\put(-35, 37){\parbox{1cm}{\raggedleft\footnotesize $40\,\unit{\tfrac{km}{h}}$}}
\put(-35, 56){\parbox{1cm}{\raggedleft\footnotesize $60\,\unit{\tfrac{km}{h}}$}}
\put(-35, 77){\parbox{1cm}{\raggedleft\footnotesize $80\,\unit{\tfrac{km}{h}}$}}
\put(52, 15){\parbox{2cm}{
\scriptsize
$\mu = +1.19\,\unit{\tfrac{km}{h}}$\newline
$\sigma = 1.45\,\unit{\tfrac{km}{h}}$\newline
$m = 1.00$\newline
}}
\end{overpic}\\[10pt]
\caption{Estimated over measured $v\subs{RL}$}%
\end{subfigure}
\hfill
\begin{subfigure}{0.49\columnwidth}
\raggedleft
\begin{overpic}[%grid
]{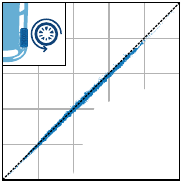}
\put(-35, -10){\parbox{1cm}{\raggedleft\footnotesize $0\,\unit{\tfrac{km}{h}}$}}
\put(20, -10){\parbox{1cm}{\raggedleft\footnotesize $40\,\unit{\tfrac{km}{h}}$}}
\put(60, -10){\parbox{1cm}{\raggedleft\footnotesize $80\,\unit{\tfrac{km}{h}}$}}
\put(-35, 18){\parbox{1cm}{\raggedleft\footnotesize $20\,\unit{\tfrac{km}{h}}$}}
\put(-35, 37){\parbox{1cm}{\raggedleft\footnotesize $40\,\unit{\tfrac{km}{h}}$}}
\put(-35, 56){\parbox{1cm}{\raggedleft\footnotesize $60\,\unit{\tfrac{km}{h}}$}}
\put(-35, 77){\parbox{1cm}{\raggedleft\footnotesize $80\,\unit{\tfrac{km}{h}}$}}
\put(52, 15){\parbox{2cm}{
\scriptsize
$\mu = +0.12\,\unit{\tfrac{km}{h}}$\newline
$\sigma = 0.73\,\unit{\tfrac{km}{h}}$\newline
$m = 0.97$\newline
}}
\end{overpic}\\[10pt]
\caption{Estimated over measured $v\subs{RR}$}%
\end{subfigure}\\[-2pt]

\begin{subfigure}{0.49\columnwidth}
\raggedleft
\begin{overpic}[%grid
]{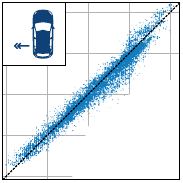}
\put(-35, -10){\parbox{1cm}{\raggedleft\footnotesize $-6\,\unit{\tfrac{m}{s^2}}$}}
\put(3, -10){\parbox{1cm}{\raggedleft\footnotesize $-4\,\unit{\tfrac{m}{s^2}}$}}
\put(48, -10){\parbox{1cm}{\raggedleft\footnotesize $0\,\unit{\tfrac{m}{s^2}}$}}
%
%\put(-35, -1){\parbox{1cm}{\raggedleft\footnotesize $-6\,\unit{\tfrac{m}{s^2}}$}}
\put(-35, 22){\parbox{1cm}{\raggedleft\footnotesize $-4\,\unit{\tfrac{m}{s^2}}$}}
\put(-35, 45){\parbox{1cm}{\raggedleft\footnotesize $-2\,\unit{\tfrac{m}{s^2}}$}}
\put(-35, 67){\parbox{1cm}{\raggedleft\footnotesize $0\,\unit{\tfrac{m}{s^2}}$}}
\put(-35, 89){\parbox{1cm}{\raggedleft\footnotesize $+2\,\unit{\tfrac{m}{s^2}}$}}
\put(52, 15){\parbox{2cm}{
\scriptsize
$\mu = +0.11\,\unit{\tfrac{m}{s^2}}$\newline
$\sigma = 0.32\,\unit{\tfrac{m}{s^2}}$\newline
$m = 0.96$\newline
}}
\end{overpic}\\[10pt]
\caption{Correlation of estimated $a\subs{lat}$ over measured $a\subs{lat}$. The visible vertical columns stem from the discretization of the vehicle's accelerometer.\newline\phantom{.}}%
\end{subfigure}
\hfill
%\begin{subfigure}{0.49\columnwidth}
%\raggedleft
%\begin{overpic}[%grid
%]{images/c2-kreuz-korr-ii-3-VR}
%\put(-35, -10){\parbox{1cm}{\raggedleft\footnotesize $0\,\unit{\tfrac{km}{h}}$}}
%\put(20, -10){\parbox{1cm}{\raggedleft\footnotesize $40\,\unit{\tfrac{km}{h}}$}}
%\put(60, -10){\parbox{1cm}{\raggedleft\footnotesize $80\,\unit{\tfrac{km}{h}}$}}
%%
%\put(-35, 18){\parbox{1cm}{\raggedleft\footnotesize $20\,\unit{\tfrac{km}{h}}$}}
%\put(-35, 37){\parbox{1cm}{\raggedleft\footnotesize $40\,\unit{\tfrac{km}{h}}$}}
%\put(-35, 56){\parbox{1cm}{\raggedleft\footnotesize $60\,\unit{\tfrac{km}{h}}$}}
%\put(-35, 77){\parbox{1cm}{\raggedleft\footnotesize $80\,\unit{\tfrac{km}{h}}$}}
%%
%%\put(52, 15){\parbox{2cm}{
%%\scriptsize
%%$\mu = +0.12\,\unit{\tfrac{km}{h}}$\newline
%%$\sigma = 0.73\,\unit{\tfrac{km}{h}}$\newline
%%$m = 0.97$\newline
%%}}
%\end{overpic}\\[10pt]
%\caption{Errors from $\delta_{C^2}, v_{C^2}$ over $s$}%
%\end{subfigure}\\[10pt]
\begin{subfigure}{0.49\columnwidth}
\raggedleft
\begin{overpic}[%grid
]{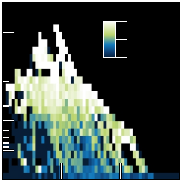}
\put(-13, -10){\parbox{1cm}{\raggedleft\footnotesize $20\,\unit{\tfrac{km}{h}}$}}
\put(20, -10){\parbox{1cm}{\raggedleft\footnotesize $40\,\unit{\tfrac{km}{h}}$}}
\put(50, -10){\parbox{1cm}{\raggedleft\footnotesize $60\,\unit{\tfrac{km}{h}}$}}
\put(66, -10){\parbox{1cm}{\raggedleft\footnotesize $v\subs{lon}$}}
\put(-35, -1){\parbox{1cm}{\raggedleft\footnotesize $0$}}
\put(-35, 13){\parbox{1cm}{\raggedleft\footnotesize $\nicefrac{1}{100\,\unit{m}}$}}
\put(-35, 30){\parbox{1cm}{\raggedleft\footnotesize $\nicefrac{1}{50\,\unit{m}}$}}
\put(-20, 50){\parbox{1cm}{\footnotesize $|\kappa|$}}
\put(-35, 80){\parbox{1cm}{\raggedleft\footnotesize $\nicefrac{1}{20\,\unit{m}}$}}
\put(75, 66){\parbox{1cm}{\color{white}\scriptsize $0\%$}}
\put(75, 76){\parbox{1cm}{\color{white}\scriptsize $50\%$}}
\put(75, 86){\parbox{1cm}{\color{white}\scriptsize $100\%$}}
\end{overpic}\\[10pt]
\caption{Percentage of cases where $v\subs{RL}$ is underestimated by more than $3\,\%$, over the two parameters absolute curvature and speed; curvatures relate to the radii in Fig.~\ref{subfig:interchange-correlation-percentage-location-RL-radii}.}%
\label{subfig:interchange-correlation-percentage}
\end{subfigure}%\\[10pt]

%\begin{subfigure}{0.49\columnwidth}
%\raggedleft
%\begin{overpic}[%grid
%]{images/c2-kreuz-korr-ii-4-RL-1A}
%\put(16, 49){\parbox{1cm}{\scriptsize $0\%$}}
%\put(16, 59){\parbox{1cm}{\scriptsize $50\%$}}
%\put(16, 69){\parbox{1cm}{\scriptsize $100\%$}}
%\end{overpic}
%\caption{Percentage by location of cases where $v\subs{RL}$ is underestimated by more than $3\,\%$. In each sharp turn (radii in $\unit{m}$ indicated), most measurements underestimate $v\subs{RL}$.}%
%\end{subfigure}
%\hfill
%\begin{subfigure}{0.49\columnwidth}
%\raggedleft
%\begin{overpic}[%grid
%]{images/c2-kreuz-korr-ii-4-RR-1A}
%\put(16, 49){\parbox{1cm}{\scriptsize $0\%$}}
%\put(16, 59){\parbox{1cm}{\scriptsize $50\%$}}
%\put(16, 69){\parbox{1cm}{\scriptsize $100\%$}}
%\end{overpic}
%\caption{Percentage by location of cases where $v\subs{RR}$ is underestimated by more than $3\,\%$. Note that, unlike $v\subs{RL}$, $v\subs{RR}$ is rarely underestimated by so much during sharp turns. }%
%\end{subfigure}

\caption{Accuracy of parameters at the interchange data set. Most parameters vary within the expected range of position measurement uncertainties, with the exception of systematically underestimated $v\subs{FL}$, $v\subs{RL}$ (indicated by arrows). Analysis in (h) and in Fig.~\ref{fig:interchange-correlation-location} indicates that strong errors in the latter two parameters primarily correlate with location, rather than with speed or lateral acceleration.}%
\label{fig:interchange-correlation}%
\vspace{-13pt}
\end{figure}

\subsection{Highway Interchange}\label{sec:tests-interchange}

While allowing for extreme maneuvers that are impossible on public roads, conditions on the closed-off test track are benign due to its very planar, homogeneous surface with relatively high friction coefficients. To evaluate the model under realistic conditions, test drives were conducted with the same car on a highway interchange (Fig.~\ref{fig:interchange-aer}) within the TAF-BW test area south of Karlsruhe, Germany, between roads K9657 and L605 ($48^\circ 59'30''\unit{N}, 8^\circ 23'\unit{E}$, cf. Fig.~\ref{fig:interchange-aer}) with the RTK GPS base station placed on the overpass. One hour of driving data was collected on and near the interchange with a setup identical to Sec.~\ref{sec:tests-campusost}, but due to occlusions of the RTK GPS signal under the overpass, regions with unreliable data were removed from the set (cf. Fig.~\ref{fig:interchange-quality}), leaving about 28\,min of significant data.

The driving style was chosen to be rather dynamic (depending on traffic), with longitudinal accelerations of $a\subs{lon}$ within $[-4.5\,\unit{m/s^2}, 3.2\,\unit{m/s^2}]$ and lateral accelerations $|a\subs{lat}|$ of up to $6\,\unit{m/s^2}$, at speed limits between $50\,\unit{km/h}$ and $80\,\unit{km/h}$.

\begin{figure}%
\begin{subfigure}{\columnwidth}
\raggedleft
\begin{overpic}[%grid
]{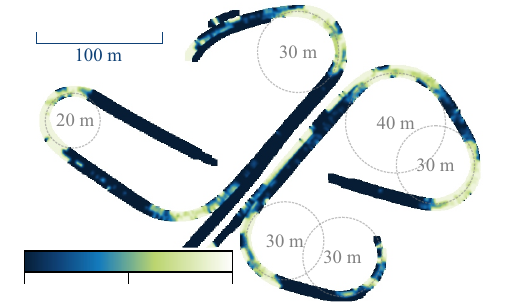}
\put(4, 0){\footnotesize $0\%$}
\put(23, 0){\footnotesize $50\%$}
\put(43, 0){\footnotesize $100\%$}
\end{overpic}
\caption{Percentage of cases where $v\subs{RL}$ is underestimated by more than $3\,\%$, with turn radii.}%
\label{subfig:interchange-correlation-percentage-location-RL-radii}
\end{subfigure}\\[10pt]

\begin{subfigure}{\columnwidth}
\raggedleft
\begin{overpic}[%grid
]{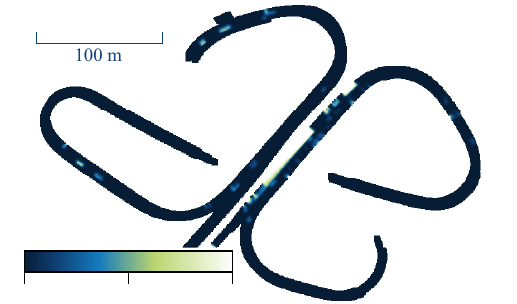}
\put(4, 0){\footnotesize $0\%$}
\put(23, 0){\footnotesize $50\%$}
\put(43, 0){\footnotesize $100\%$}
\end{overpic}
\caption{Percentage of cases where $v\subs{RR}$ is underestimated by more than $3\,\%$.}%
\end{subfigure}

\caption{Accuracy of parameters at the interchange data set (continued from Fig.~\ref{fig:interchange-correlation}): Percentage by location of cases where $v\subs{RL}$ (a) or $v\subs{RR}$ (b) is underestimated by more than $3\,\%$. In each sharp turn, most measurements underestimate $v\subs{RL}$ (a) regardless of speed (cf. Fig.~\ref{subfig:interchange-correlation-percentage}, while $v\subs{RR}$ is rarely underestimated by so much in the same instances.}%
\label{fig:interchange-correlation-location}%
\end{figure}

\subsubsection{Evaluation of the Analytic Model}

Application of the model shows overall good accuracy consistent with the results from Sec.~\ref{sec:tests-campusost}: A typical extract of $\delta\subs{SWA}$ and $v\subs{lon}$ is shown in Fig.~\ref{fig:interchange-quality-delta-v}; average errors in the estimation of $\delta\subs{SWA}$ are shown in Fig.~\ref{fig:interchange-quality-delta-errors}: Errors of up to $10^\circ$ (or $10\,\%$) occurred near sharp turns, but mean and median errors were at $2.76^\circ$ and $1.89^\circ$ respectively. The distribution shows most errors before and after the turn, not during the turn, which hints at synchronization errors between the RTK GPS data and the CAN bus messages.

The scatter plots in Fig.~\ref{fig:interchange-correlation} show an increase in \emph{absolute} errors compared to the test drives on the closed-off test track, but a slight decrease in errors \emph{relative} to the vehicle speed. A notable exception is the left-side speeds $v\subs{FL}$ (Fig.~\ref{fig:interchange-correlation}c) and $v\subs{RL}$ (Fig.~\ref{fig:interchange-correlation}e) which are visibly underestimated near $40\,\unit{km/h}$. Analysis in the parameters (Fig.~\ref{fig:interchange-correlation}h, Fig.~\ref{fig:interchange-correlation-location}) shows that this effect occurs on both left-side wheels in all sharp turns (always the \emph{outer} wheels), but not in right-side wheels (Fig.~\ref{fig:interchange-correlation-location}).

Several model assumptions may contribute to this effect: Sharp turns can cause a loss of traction, as discussed in the previous sections, in particular on the outside wheels which bear the greater load. Also, sharp lateral accelerations can cause the vehicle to roll due to its elevated center of mass; in this case, the track width of the vehicle narrows slightly, while the GPS antenna moves to the outside. However, Fig.~\ref{fig:interchange-correlation}h shows no strong correlation of these errors with speed $v\subs{lon}$; instead, the distribution of errors is mainly governed by $\kappa$ regardless of speed. This may indicate that the ramps are significantly banked, violating the model assumption of a planar surface. Again, the measurements taken during the test drive cannot resolve this ambiguity.

\begin{figure}%
\raggedleft
\begin{overpic}[%grid
]{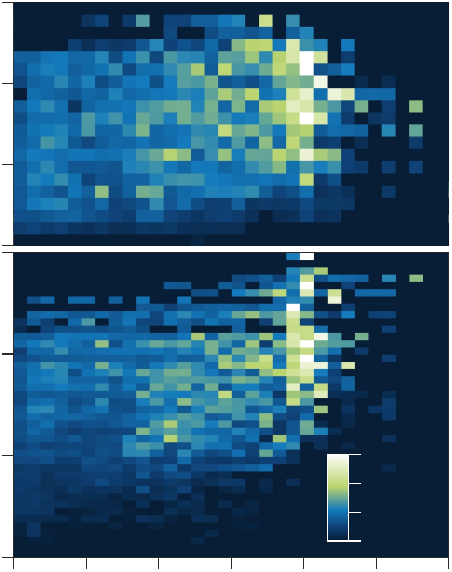}

\put(-12,1){\parbox{1cm}{\raggedleft\footnotesize $0\, \unit{km/h}$}}
\put(-12,19){\parbox{1cm}{\raggedleft\footnotesize $1\, \unit{km/h}$}}
\put(-12,28){\parbox{1cm}{\raggedleft\footnotesize $\max$\\[2pt] $|v\subs{lon}^{C^2}\!\! {-} v\subs{lon}\sups{CAN}|$}}
\put(-12,37){\parbox{1cm}{\raggedleft\footnotesize $2\, \unit{km/h}$}}
\put(-12,54){\parbox{1cm}{\raggedleft\footnotesize $3\, \unit{km/h}$}}

\put(-12,58){\parbox{1cm}{\raggedleft\footnotesize $0\, \unit{m/s^2}$}}
\put(-12,70){\parbox{1cm}{\raggedleft\footnotesize $1\, \unit{m/s^2}$}}
\put(-15,77){\parbox{1cm}{\raggedleft\footnotesize $\max |a\subs{lon}|$}}
\put(-12,84){\parbox{1cm}{\raggedleft\footnotesize $2\, \unit{m/s^2}$}}
\put(-12,98){\parbox{1cm}{\raggedleft\footnotesize $3\, \unit{m/s^2}$}}

\put(-8,-4){\parbox{1cm}{\raggedleft\footnotesize $0\, \unit{\tfrac{m}{s^2}}$}}
\put( 7,-4){\parbox{1cm}{\raggedleft\footnotesize $1\, \unit{\tfrac{m}{s^2}}$}}
\put(20,-4){\parbox{1cm}{\raggedleft\footnotesize $2\, \unit{\tfrac{m}{s^2}}$}}
\put(33,-4){\parbox{1cm}{\raggedleft\footnotesize $3\, \unit{\tfrac{m}{s^2}}$}}
\put(33,-8){\parbox{1cm}{\raggedleft\footnotesize $\max |a\subs{lat}|$}}
\put(45,-4){\parbox{1cm}{\raggedleft\footnotesize $4\, \unit{\tfrac{m}{s^2}}$}}
\put(59,-4){\parbox{1cm}{\raggedleft\footnotesize $5\, \unit{\tfrac{m}{s^2}}$}}
\put(71,-4){\parbox{1cm}{\raggedleft\footnotesize $6\, \unit{\tfrac{m}{s^2}}$}}

\put(65,5){\parbox{1cm}{\color{white}\raggedright\footnotesize $1\, \unit{m}$}}
\put(65,10){\parbox{1cm}{\color{white}\raggedright\footnotesize $2\, \unit{m}$}}
\put(65,15){\parbox{1cm}{\color{white}\raggedright\footnotesize $3\, \unit{m}$}}
\put(65,20){\parbox{1cm}{\color{white}\raggedright\footnotesize $4\, \unit{m}$}}
\end{overpic}\\[19pt]
\caption{Average absolute end point error between $C^2$ forward model and RTK GPS track of connected trajectories (distances approx. within $[5\,\unit{m}, 150\,\unit{m}]$) by different parameters occurring in each trajectory. As is expected, the error increases strongly in trajectories with greater $\max |a\subs{lat}|$, while the effect of $\max |a\subs{lon}|$ is much smaller (upper plot). However the end point error scales notably with errors in (longitudinal) speed (lower plot).}%
\label{fig:extrapolations-endpointerrors}%
%\vspace{-10pt}
\end{figure}

\newcommand{\errXY}{E}

\newcommand{\mss}{\unit{\tfrac{m}{s^2}}}

\newcommand{\markerleft}[1]{\makebox[0pt][r]{$\;\;#1\;\;$}}
\newcommand{\markerright}[1]{\makebox[0pt][l]{$\;\;#1\;\;$}}

\begin{figure*}[t]%
\begin{overpic}[%grid
]{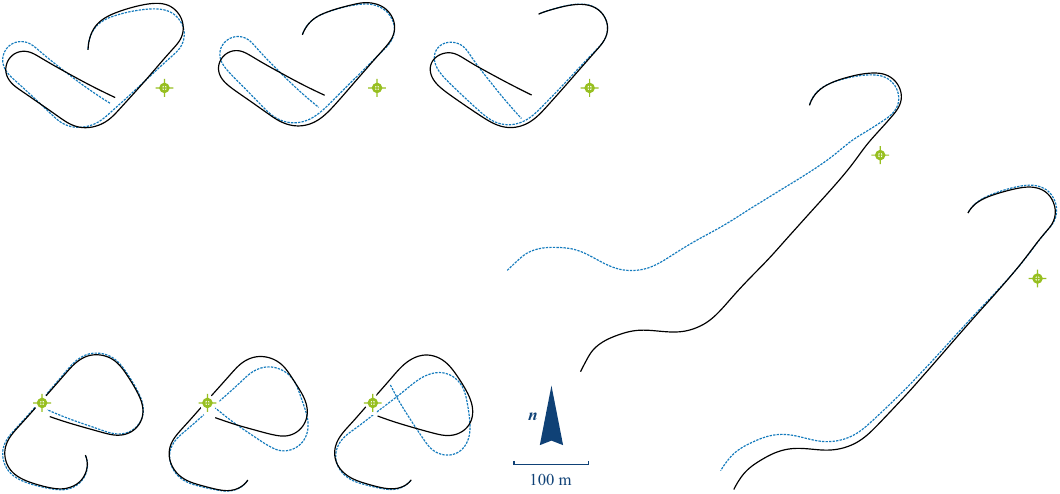}%

\put(2,15){\parbox{4cm}{\scriptsize
%6
$\langle|a\subs{lon}|\rangle \,{=}\, 0.48\mss$\;$s \,{=}\, 616\,\unit{m}$\\%[-5pt]

$\langle|a\subs{lat}|\rangle \,{=}\, 1.59\mss$\;$\errXY \,{=}\, 9.96\,\unit{m}$\\

\markerright{\blacktriangledown}
}}
\put(18,15){\parbox{4cm}{\scriptsize
%6
$\langle|a\subs{lon}|\rangle \,{=}\, 0.88\mss$\;$s \,{=}\, 580\,\unit{m}$\\%[-5pt]

$\langle|a\subs{lat}|\rangle \,{=}\, 2.25\mss$\;$\errXY \,{=}\, 13.0\,\unit{m}$\\

\markerright{\blacktriangledown}
}}
\put(34,15){\parbox{4cm}{\scriptsize
%6
$\langle|a\subs{lon}|\rangle \,{=}\, 0.86\mss$\;$s \,{=}\, 582\,\unit{m}$\\%[-5pt]

$\langle|a\subs{lat}|\rangle \,{=}\, 2.99\mss$\;$\errXY \,{=}\, 45.3\,\unit{m}$\\

\markerright{\blacktriangledown}
}}

\put(3,28.5){\parbox{4cm}{\scriptsize
%6
\markerright{\blacktriangle}\\

$\langle|a\subs{lon}|\rangle \,{=}\, 0.74\mss$\;$s \,{=}\, 696\,\unit{m}$\\%[-5pt]

$\langle|a\subs{lat}|\rangle \,{=}\, 1.45\mss$\;$\errXY \,{=}\, 10.2\,\unit{m}$
}}
\put(24,28.5){\parbox{4cm}{\scriptsize
%6
\markerright{\blacktriangle}\\

$\langle|a\subs{lon}|\rangle \,{=}\, 0.70\mss$\;$s \,{=}\, 669\,\unit{m}$\\%[-5pt]

$\langle|a\subs{lat}|\rangle \,{=}\, 1.77\mss$\;$\errXY \,{=}\, 18.0\,\unit{m}$
}}
\put(44,28.5){\parbox{4cm}{\scriptsize
%6
\markerright{\blacktriangle}\\

$\langle|a\subs{lon}|\rangle \,{=}\, 0.89\mss$\;$s \,{=}\, 618\,\unit{m}$\\%[-5pt]

$\langle|a\subs{lat}|\rangle \,{=}\, 1.94\mss$\;$\errXY \,{=}\, 34.2\,\unit{m}$
}}

\put(60,34){\parbox{4cm}{\scriptsize
%6
$\langle|a\subs{lon}|\rangle \,{=}\, 0.73\mss$\;$s \,{=}\, 767\,\unit{m}$\\[-5pt]

$\langle|a\subs{lat}|\rangle \,{=}\, 1.59\mss$\;$\errXY \,{=}\, 167\,\unit{m}$\markerright{\blacktriangleright}
}}
\put(71,14){\parbox{4cm}{\scriptsize
%6
$\;\;\langle|a\subs{lon}|\rangle \,{=}\, 0.66\mss$\;$s \,{=}\, 754\,\unit{m}$\markerright{\blacktriangleright}\\%[-5pt]

$\langle|a\subs{lat}|\rangle \,{=}\, 1.14\mss$\;$\errXY \,{=}\, 30.7\,\unit{m}$
}}

\end{overpic}
\caption{Examples of accuracy parameters for the forward model on eight connected trajectories. In each case, steering wheel angles and speeds were estimated from GPS tracks (solid, black) and then in turn used with the forward model to estimate the future trajectory. $\langle|a\subs{lon}|\rangle$ is the average absolute longitudinal acceleration, $\langle|a\subs{lat}|\rangle$ is the average absolute lateral acceleration, $s$ is the arc length of the trajectory, and $\errXY$ is the norm of the end point deviation.}%
\label{fig:extrapolations}%
\vspace{-13pt}
\end{figure*}

\subsubsection{Evaluation of the Forward Model}

The data collected at the interchange also allows for a brief evaluation on the properties of the forward model: We take the recorded $\delta\subs{SWA}\sups{CAN}$ and $v\subs{lon}\sups{CAN}$, along with the extracted steering functions from Sec.~\ref{sec:tests-campusost-swa}, and apply them to extrapolate the vehicle's path over the duration of each connected sub-trajectory (i.e. trajectories within the data set with continuous high-quality RTK GPS signal). It should be understood that this model is not meant to extrapolate from measurements, but instead to estimate the parameters during numerical optimization or simulation; however the results can serve to place the previously stated parameter estimates into a spatial context, and to compare the results herein to the simulated open-loop controller test drives in \cite{icmc}.

Figure~\ref{fig:extrapolations} shows eight typical results for connected trajectories, selected to show the effect of different degrees of dynamics onto the extrapolation accuracy. Trajectories with low average absolute accelerations exhibit visibly better extrapolation, measured by the distance between measured and extrapolated end points.

To allow for a more quantitative analysis on the increase of extrapolation error with accelerations, the data set was split into (overlapping) trajectories of length $5\,\unit{m}$ to $150\,\unit{m}$. For each trajectory, the maximal absolute acceleration (longitudinal and lateral) was determined, as well as the maximal error between estimated speed $v\subs{lon}^{C^2}$ and recorded speed $v\subs{lon}\sups{CAN}$. For these parameter combinations, a map of maximum endpoint distances is shown in Fig.~\ref{fig:extrapolations-endpointerrors}.

The results show that the end point error scales strongly with the occurring \emph{lateral} accelerations (as would be expected), but only slightly with the \emph{longitudinal} accelerations. However, it scales notably with the estimation errors in longitudinal speed. This effect was also observed in the simulations in \cite{icmc}---speed errors during turns lead to significant directional errors: A correct steering wheel angle applied for a correct time interval is applied for a wrong arc length, if the vehicle's speed is incorrect. This means that an incorrect overall turn angle is driven and the error accumulates after the turn (as seen in Fig.~\ref{fig:extrapolations}). It can be seen that in trajectories with very little $\max |a\subs{lat}|$ (leftmost column), the effect of speed errors $\max |v\subs{lon}^{C^2}\!\! {-} v\subs{lon}\sups{CAN}|$ is considerably lower, because such trajectories do not contain significant turns.

\section{Conclusion and Outlook}\label{sec:conclusion}

We have evaluated a model for vehicle dynamics \cite{icmc} that is particularly convenient for maneuver planning applications, due to its lightweight computation and compatibility with both iterative optimization and generative forward integration from given control parameters, which is desirable if both methods are combined during planning. Real-world test drives were conducted both on a public road, and on a closed-off test track (for more uncommon dynamics), to provide an estimate of systematic errors by various violations of the model's assumptions in typical road traffic situations, and in two different applications: Using the model to analyze a given trajectory for its underlying control inputs, e.g. for implementing hard or soft constraints during trajectory optimization or estimating internal states from observations; and to predict a vehicle's path given its control inputs. The results indicate that while the model can be sufficient for a range of \emph{analytic} purposes, its \emph{predictive} accuracy is limited even in moderately dynamic driving situations.

%The results indicate that the model accuracy is consistent with the results of the previous simulation experiments (although the observable quality is limited by measurement uncertainties) in that the model achieves good accuracy while the vehicle operates close to the model assumptions, but deviates when these assumptions are violated. Realistic quantitative estimates of the relevant effects are provided.

\subsection*{Outlook}

A clear specification of the adequacy of a kinematic vehicle dynamics model under different conditions is clearly yet to be established: This adequacy depends on the purpose as well as on the available data and computational power. Model applications to traffic observation thus need to account for uncertain vehicle parameters, while planning and prediction applications must assume that ``high definition (HD) maps'' of road conditions and road geometry may not be available.

Therefore, a more exhaustive evaluation has to consider different vehicle types (including trucks or coaches with multiple steered or unsteered axles) as well as a systematic measurement of road and vehicle properties, such that accurate parameters for more complex models can be provided and the models can be directly related in terms of model accuracy and demand for input parameters and computational power.

Furthermore, extensions of the discussed model can allow to consider parameters such as road surface curvature, without substantially increasing demand for known parameters or computational power. This may provide another intermediate option between simple planar kinematic models, and complex dynamic vehicle models.

\bibliographystyle{abbrv}
\bibliography{rootC2}

\end{document}

%% file: includes.tex
\newcommand{\traj}[0]{\xi}

\newcommand{\dee}[0]{\mathrm{d}}

\newcommand{\subs}[1]{_{\text{#1}}}
\newcommand{\sups}[1]{^{\text{#1}}}
\newcommand{\transp}{^{\mathsf T}}